\documentclass[letterpaper, 10 pt, conference]{ieeeconf}

\IEEEoverridecommandlockouts                              % This command is only needed if 
\usepackage{graphics} % for pdf, bitmapped graphics files
\usepackage{epsfig} % for postscript graphics files
\usepackage{mathptmx} % assumes new font selection scheme installed
\usepackage{times} % assumes new font selection scheme installed
\usepackage{amsmath} % assumes amsmath package installed
\usepackage{amssymb}  % assumes amsmath package installed
\usepackage{url}
\usepackage{booktabs}
\usepackage{siunitx}
\usepackage{csquotes}
\usepackage{pgf}
\usepackage{pgfplots}
\usepackage[T1]{fontenc}

\usepackage{multirow} 
\usepackage{pifont} % for the checkmark and x symbol in the tables
\usepackage{arydshln} % for dashed lines in tables
\usepackage{subfiles}
\usepackage{balance} % balance columns on last page
\usepackage[absolute]{textpos} % used for the review and IEEE copyright notice
\usepackage{xcolor}
\usepackage{graphicx} % for teaser gallery
\usepackage{subcaption} % for subfigures

\usepackage{pifont}% http://ctan.org/pkg/pifont
	\newcommand{\cmark}{\ding{51}}%
	\newcommand{\xmark}{\ding{55}}%

\newcolumntype{x}[1]{>{\centering\arraybackslash\hspace{0pt}}p{#1}}

\usepackage{caption}
\usepackage{etoolbox}
\makeatletter
\patchcmd{\@makecaption}
  {\scshape}
  {}
  {}
  {}
\makeatletter
\patchcmd{\@makecaption}
  {\\}
  {.\ }
  {}
  {}
\makeatother

\title{Survey on Datasets for Perception in Unstructured Outdoor Environments}
\let\oldtwocolumn\twocolumn
\renewcommand\twocolumn[1][]{%
	\oldtwocolumn[{#1}{
\definecolor{color1}{RGB}{255, 0, 0}
\definecolor{color2}{RGB}{255, 38, 0}
\definecolor{color3}{RGB}{255, 75, 0}
\definecolor{color4}{RGB}{255, 113, 0}
\definecolor{color5}{RGB}{255, 150, 0}
\definecolor{color6}{RGB}{255, 188, 0}
\definecolor{color7}{RGB}{255, 225, 0}
\definecolor{color8}{RGB}{239, 247, 0}
\definecolor{color9}{RGB}{207, 254, 0}
\definecolor{color10}{RGB}{170, 255, 0}
\definecolor{color11}{RGB}{133, 255, 0}
\definecolor{color12}{RGB}{95, 255, 0}
\definecolor{color13}{RGB}{58, 255, 0}
\definecolor{color14}{RGB}{20, 255, 0}
\definecolor{color15}{RGB}{8, 255, 25}
\definecolor{color16}{RGB}{0, 255, 54}
\definecolor{color17}{RGB}{0, 255, 92}
\definecolor{color18}{RGB}{0, 255, 130}
\definecolor{color19}{RGB}{0, 255, 168}
\definecolor{color20}{RGB}{0, 255, 205}
\definecolor{color21}{RGB}{0, 245, 234}
\definecolor{color22}{RGB}{0, 226, 253}
\definecolor{color23}{RGB}{0, 191, 255}
\definecolor{color24}{RGB}{0, 154, 255}
\definecolor{color25}{RGB}{0, 116, 255}
\definecolor{color26}{RGB}{0, 78, 255}
\definecolor{color27}{RGB}{0, 40, 255}
\definecolor{color28}{RGB}{12, 15, 255}
\definecolor{color29}{RGB}{34, 0, 255}
\definecolor{color30}{RGB}{72, 0, 255}
\definecolor{color31}{RGB}{109, 0, 255}
\definecolor{color32}{RGB}{147, 0, 255}
\definecolor{color33}{RGB}{185, 0, 255}
\definecolor{color34}{RGB}{220, 0, 252}
\definecolor{color35}{RGB}{251, 0, 245}
\definecolor{color36}{RGB}{255, 0, 211}
\definecolor{color37}{RGB}{255, 0, 174}
\definecolor{color38}{RGB}{255, 0, 136}
\definecolor{color39}{RGB}{255, 0, 99}
\definecolor{color40}{RGB}{255, 0, 61}
		\newlength{\teaserimageheight}
		\setlength{\teaserimageheight}{4.em} % Set the height you want for your images
		\begin{center}
            \centering
            \fboxsep=0pt%padding thickness
            \fboxrule=2pt%border thickness	
            \fcolorbox{color1}{white}
            {\href{https://github.com/ulaval-damas/tree-bark-classification?tab=readme-ov-file\#barknet-10-database}{\includegraphics[height=\teaserimageheight]{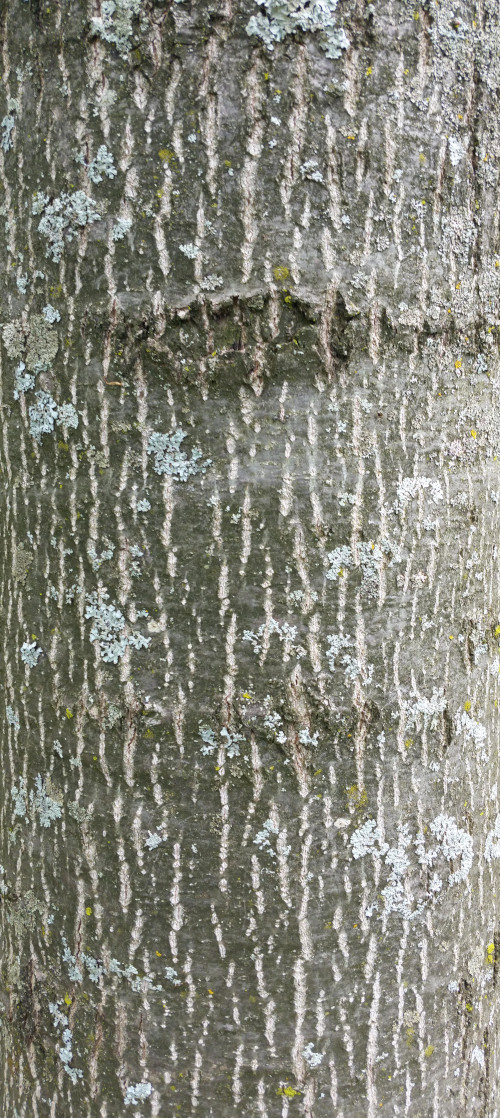}}}
            \fcolorbox{color2}{white}
            {\href{https://github.com/ulaval-damas/tree-bark-classification?tab=readme-ov-file\#barknet-10-database}{\includegraphics[height=\teaserimageheight]{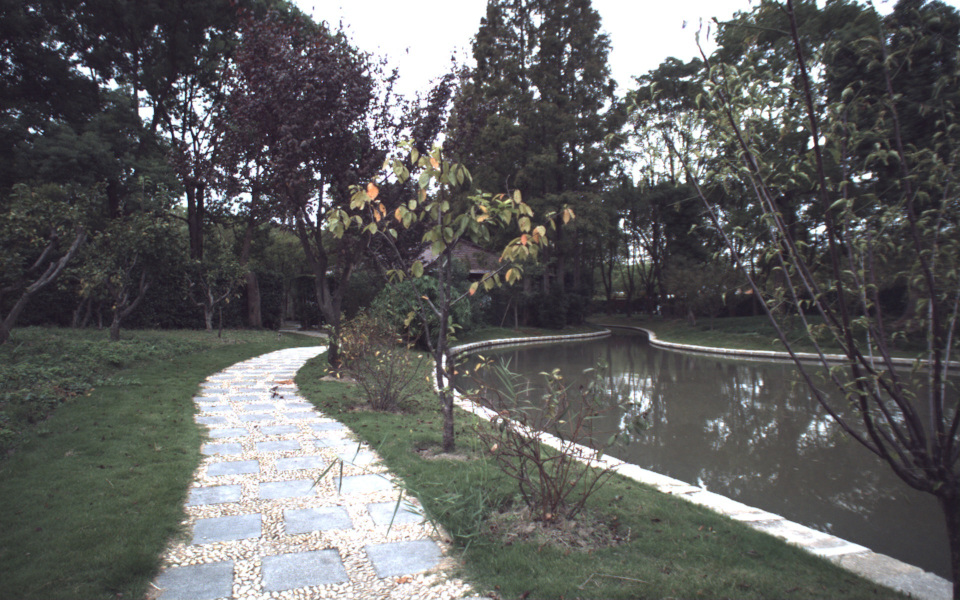}}}
		\fcolorbox{color3}{white}
            {\href{https://github.com/norlab-ulaval/PercepTreeV1?tab=readme-ov-file\#Datasets}{\includegraphics[height=\teaserimageheight]{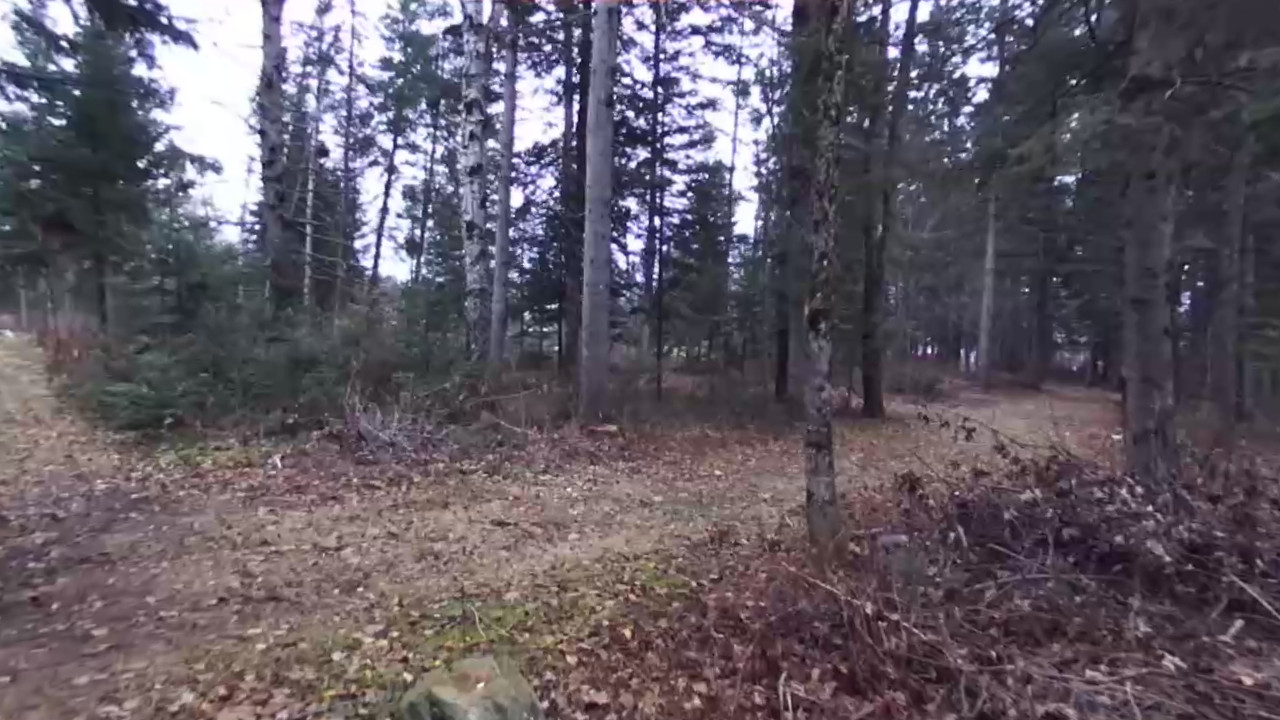}}}
            \fcolorbox{color4}{white}
            {\href{https://www.cavs.msstate.edu/resources/autonomous_dataset.php}{\includegraphics[height=\teaserimageheight]{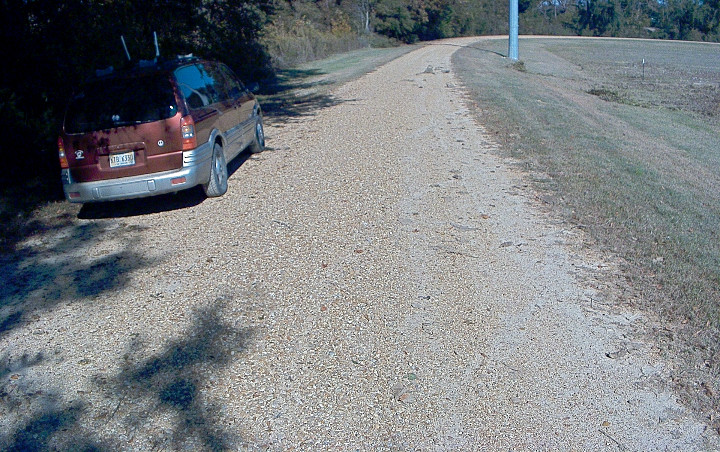}}}
            \fcolorbox{color5}{white}
            {\href{https://www.cavs.msstate.edu/resources/autonomous_dataset.php}{\includegraphics[height=\teaserimageheight]{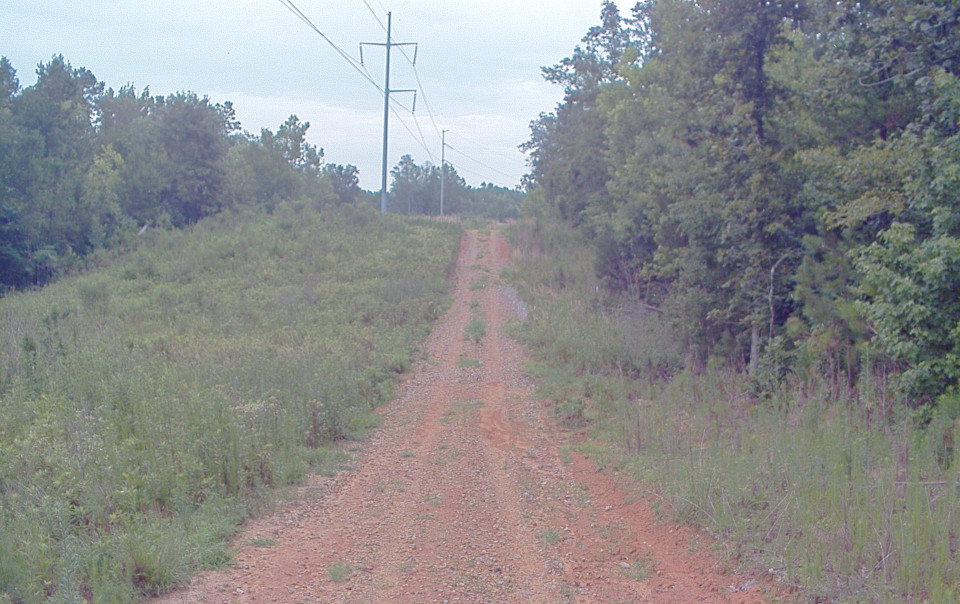}}}
            \fcolorbox{color6}{white}
            {\href{https://lhoangan.github.io/eden/}{\includegraphics[height=\teaserimageheight]{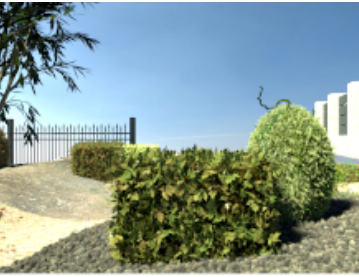}}}
            \fcolorbox{color7}{white}
            {\href{https://github.com/juanb09111/FinnForest}{\includegraphics[height=\teaserimageheight]{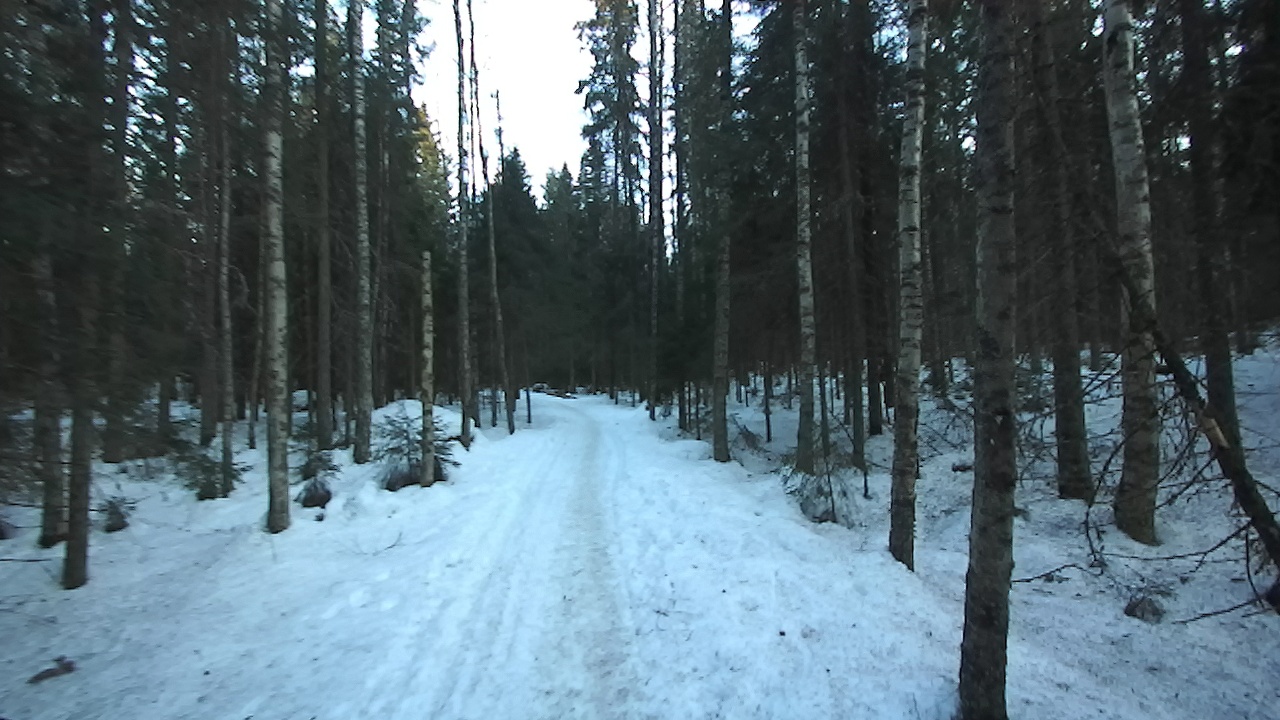}}}
            \fcolorbox{color8}{white}
            {\href{https://zenodo.org/records/5213825}{\includegraphics[height=\teaserimageheight]{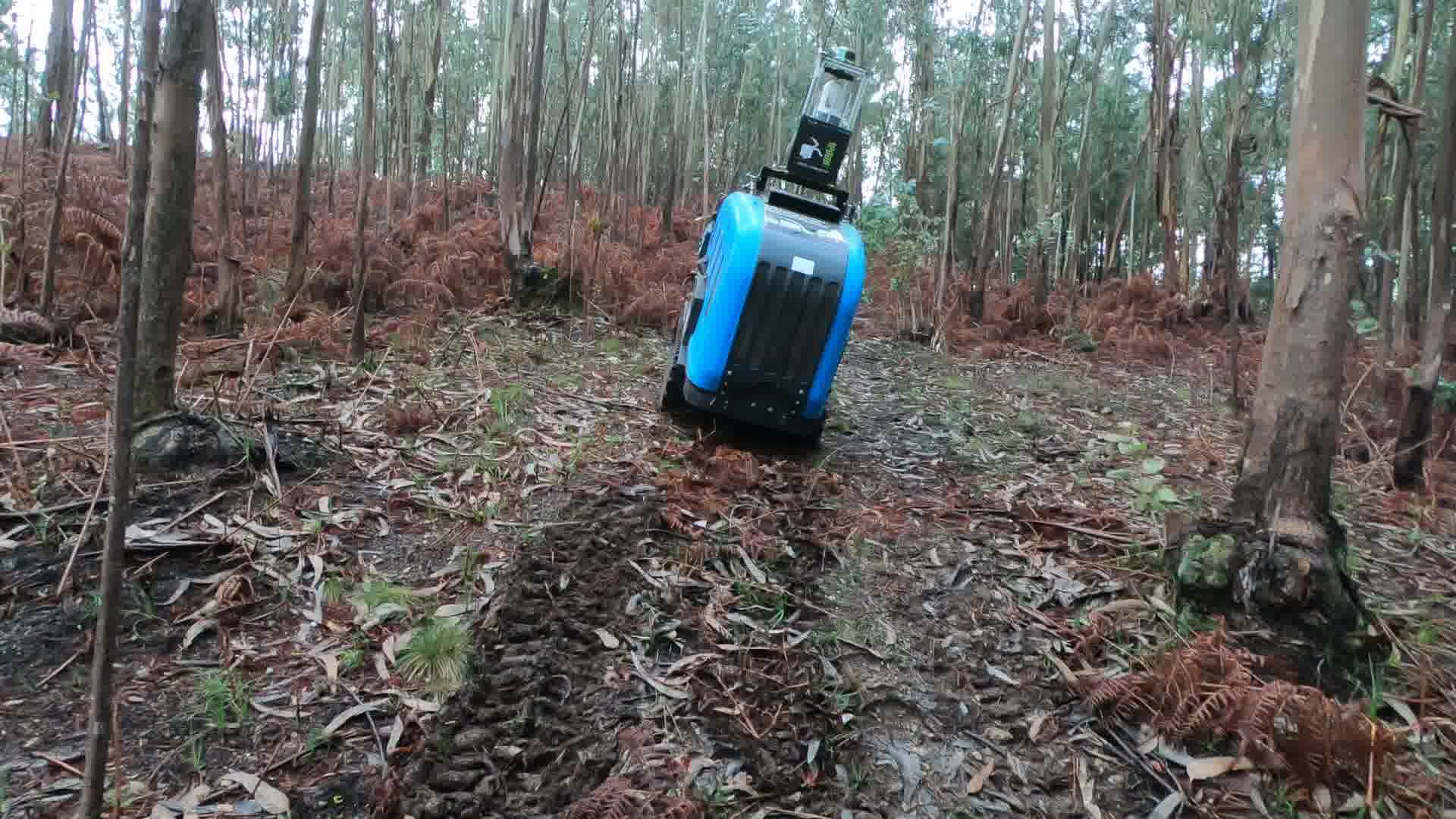}}}
		\fcolorbox{color9}{white}
            {\href{http://deepscene.cs.uni-freiburg.de\#datasets}{\includegraphics[height=\teaserimageheight]{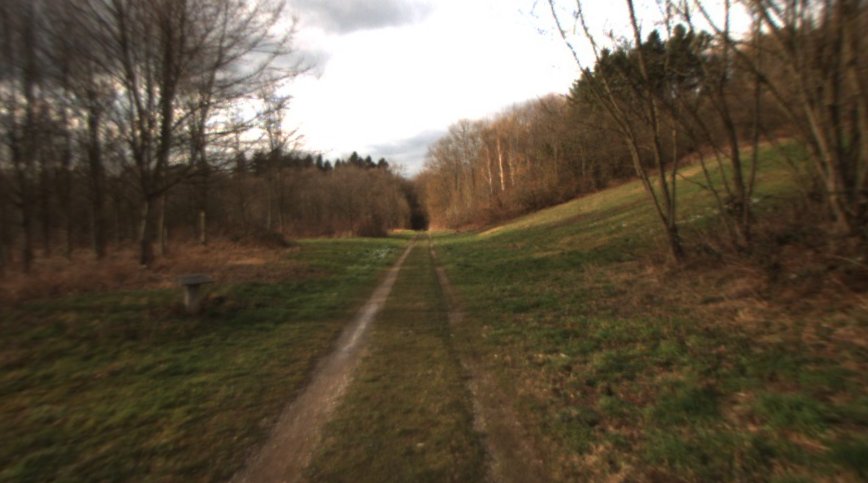}}}
            \fcolorbox{color10}{white}
            {\href{https://goose-dataset.de/}{\includegraphics[height=\teaserimageheight]{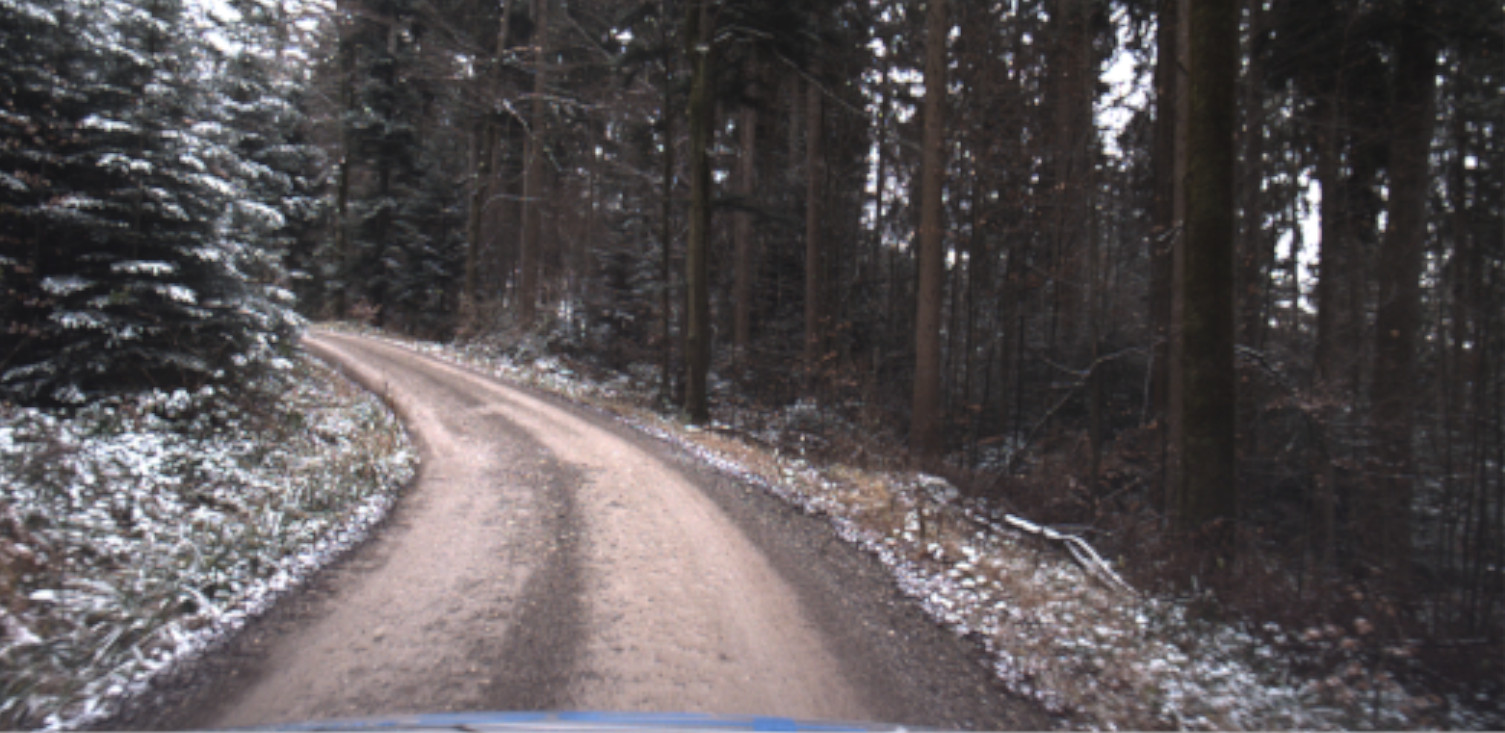}}}
            \fcolorbox{color11}{white}
            {\href{https://osf.io/pku3e/}{\includegraphics[height=\teaserimageheight]{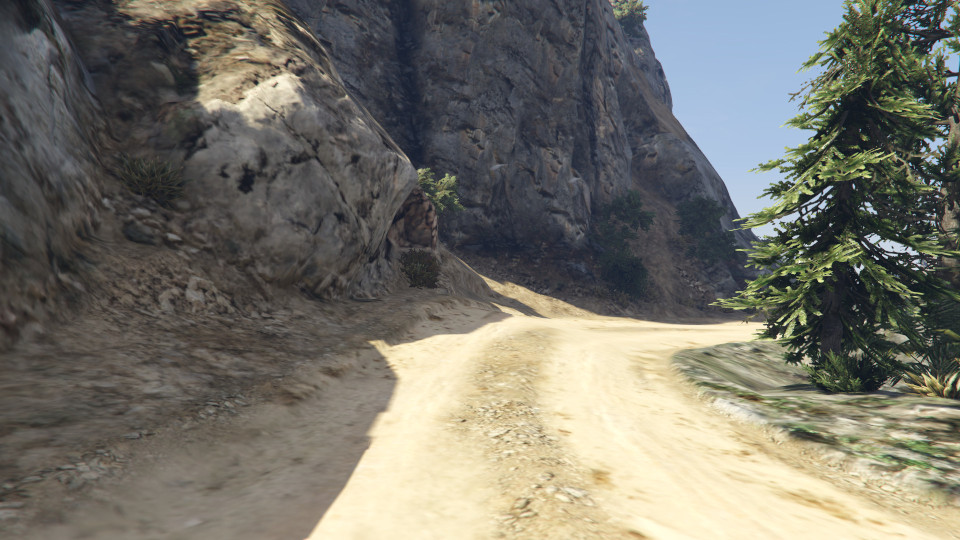}}}
		\fcolorbox{color12}{white}
            {\href{https://www.cavs.msstate.edu/resources/autonomous_dataset.php}{\includegraphics[height=\teaserimageheight]{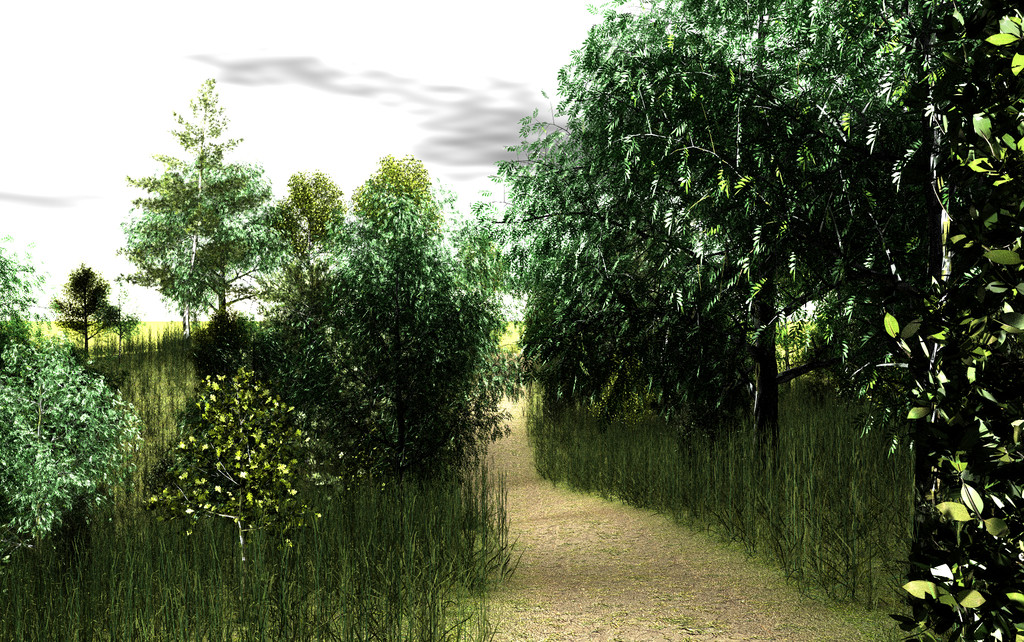}}}
            \fcolorbox{color13}{white}
            {\href{https://norlab.ulaval.ca/research/montmorencydataset/}{\includegraphics[height=\teaserimageheight]{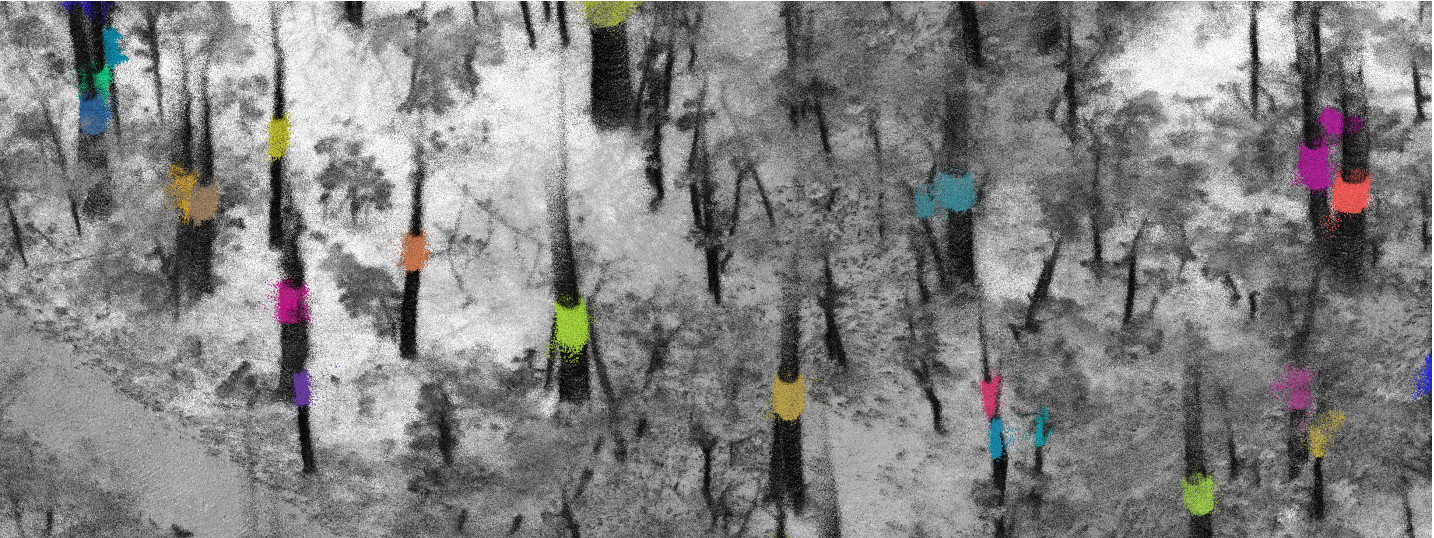}}}
		\fcolorbox{color14}{white}
            {\href{https://www.dfki.uni-kl.de/~neigel/offsed.html}{\includegraphics[height=\teaserimageheight]{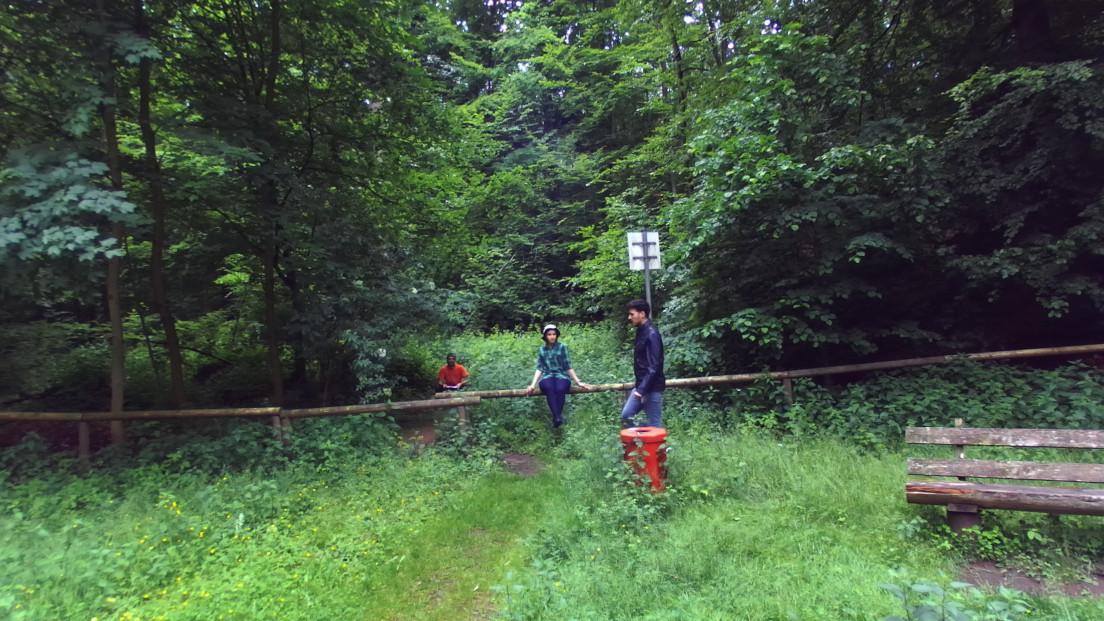}}}
		\fcolorbox{color15}{white}
            {\href{https://github.com/chaytonmin/Off-Road-Freespace-Detection\#Prepare-data}{\includegraphics[height=\teaserimageheight]{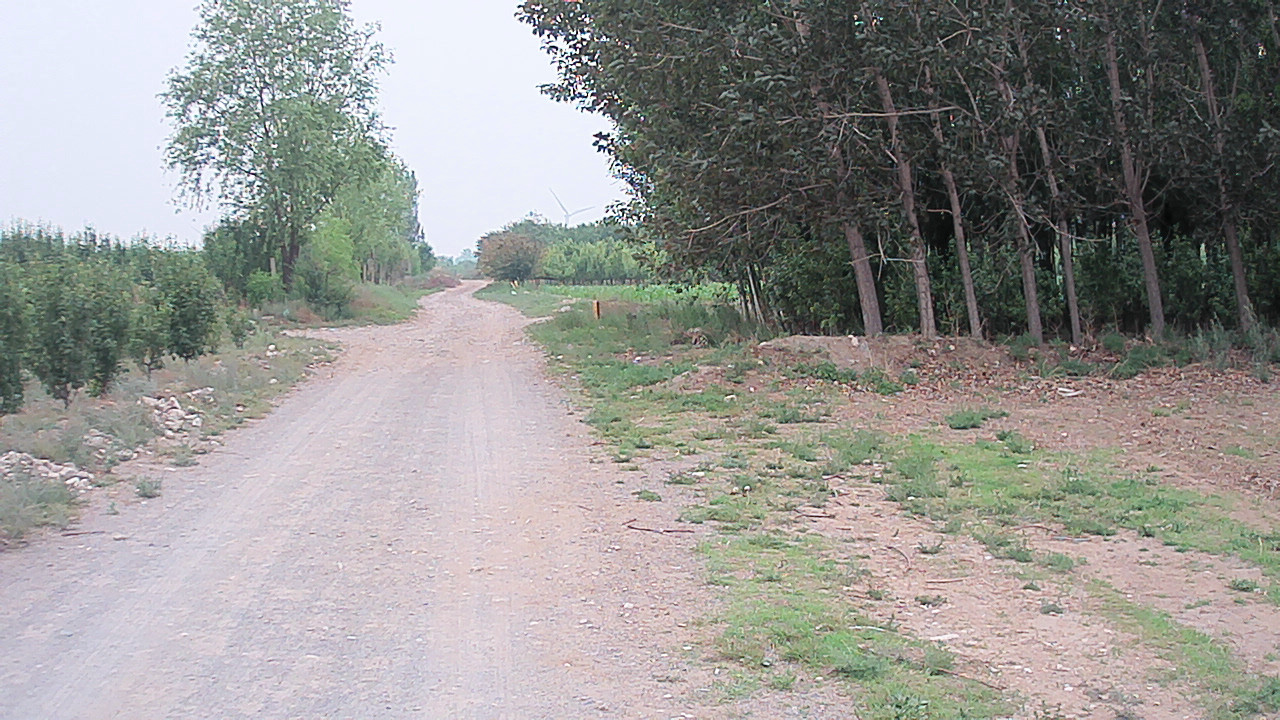}}}
            \fcolorbox{color16}{white}
            {\href{https://github.com/UT-ADL/e2e-rally-estonia\#dataset}{\includegraphics[height=\teaserimageheight]{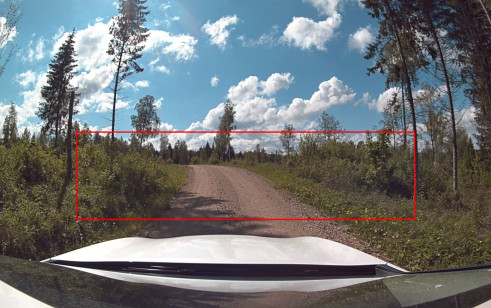}}}
            \fcolorbox{color17}{white}
            {\href{https://github.com/unmannedlab/RELLIS-3D}{\includegraphics[height=\teaserimageheight]
                {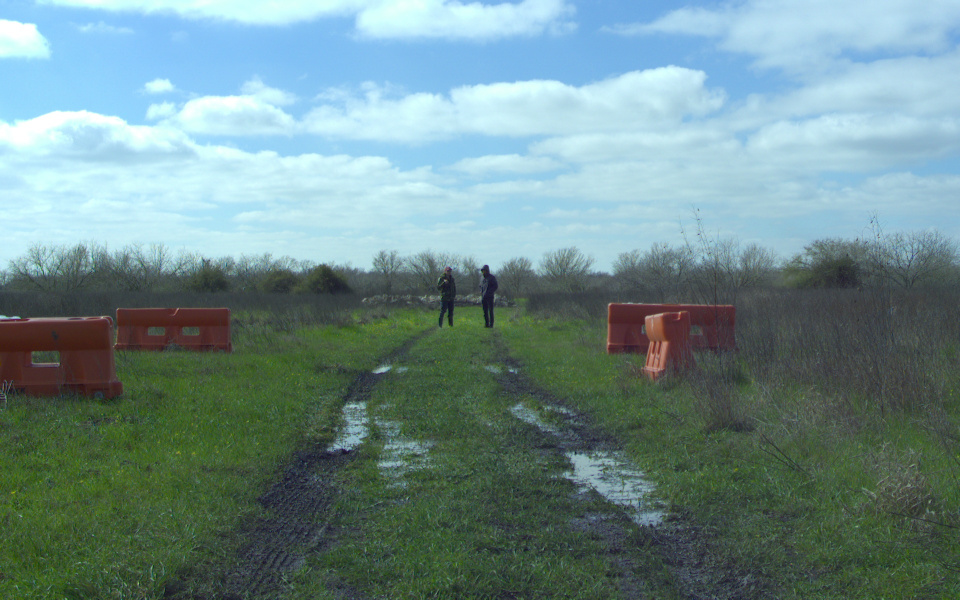}}}
		\fcolorbox{color18}{white}{\href{http://rugd.vision/}
            {\includegraphics[height=\teaserimageheight]{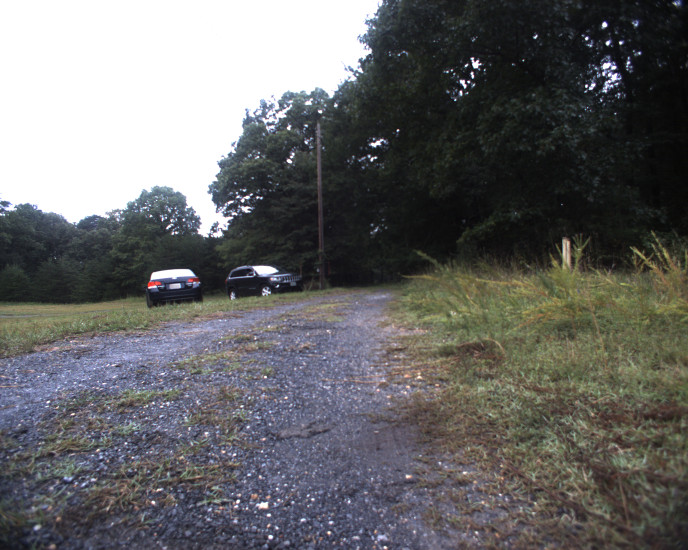}}}
            \fcolorbox{color19}{white}{\includegraphics[height=\teaserimageheight]{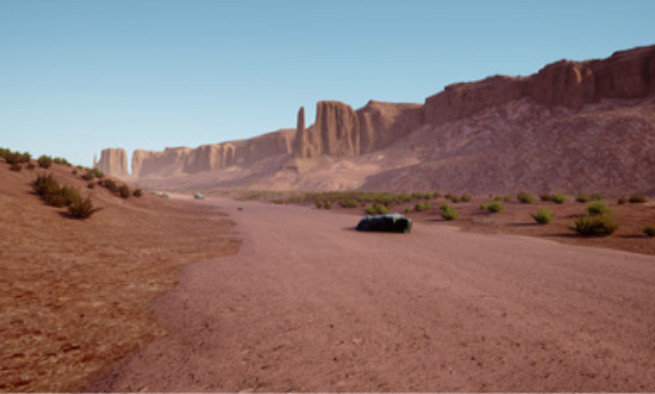}}
            \fcolorbox{color20}{white}{\href{https://zenodo.org/records/6369446}
            {\includegraphics[height=\teaserimageheight]{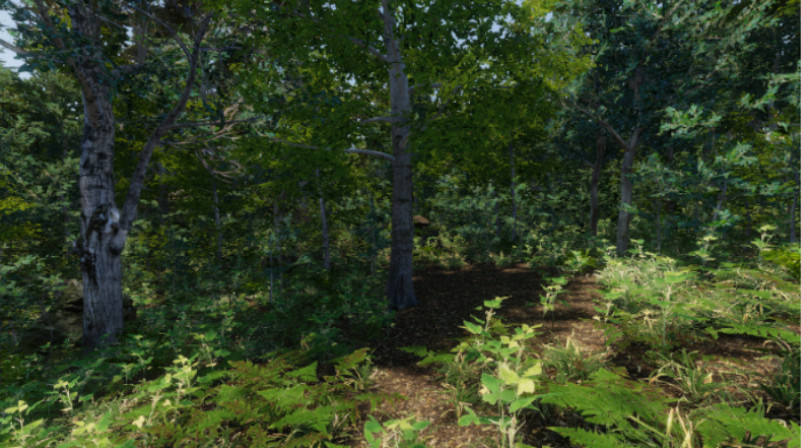}}}
		\fcolorbox{color21}{white}
            {\href{https://github.com/norlab-ulaval/PercepTreeV1?tab=readme-ov-file\#Datasets}{\includegraphics[height=\teaserimageheight]{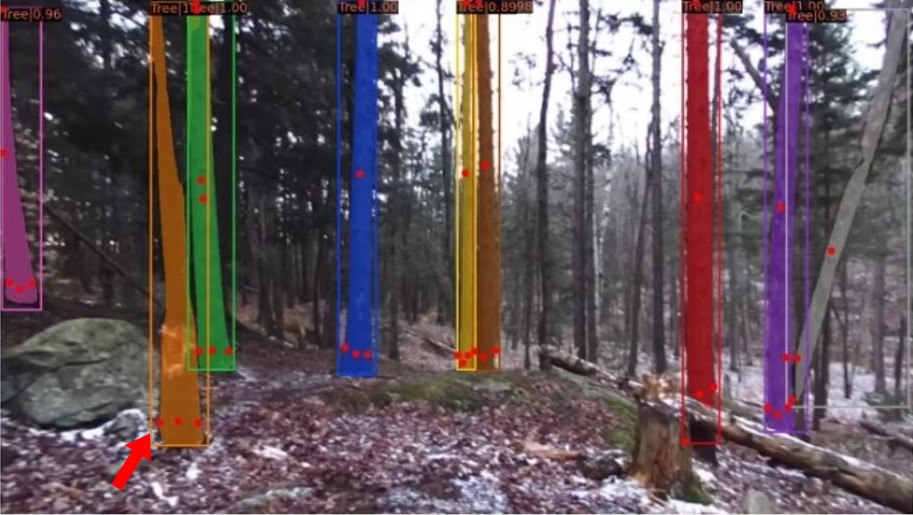}}}
		\fcolorbox{color22}{white}
            {\href{https://theairlab.org/TartanDrive2/}{\includegraphics[height=\teaserimageheight]{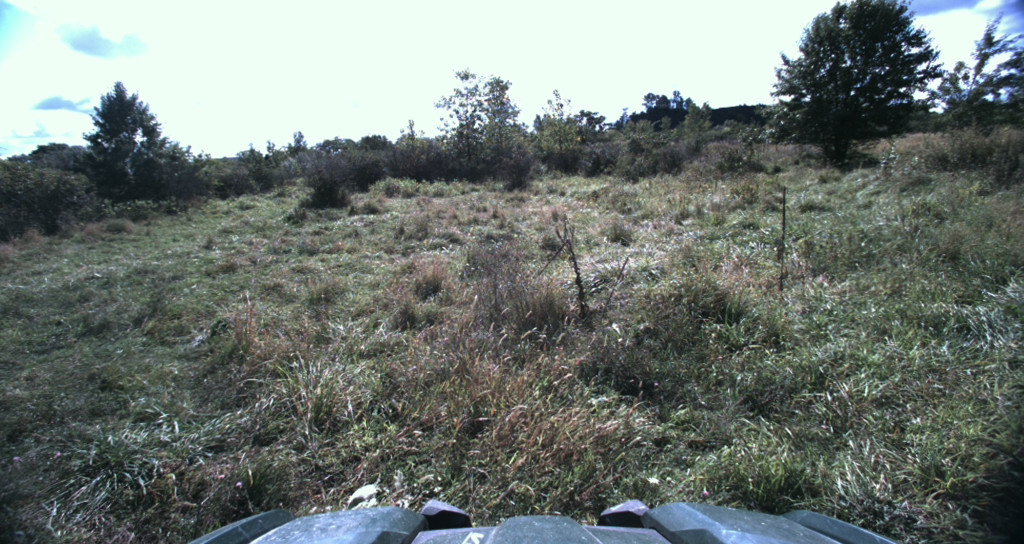}}}
            \fcolorbox{color23}{white}
            {\href{https://mucar3.de/icpr2020-tas500/}{\includegraphics[height=\teaserimageheight]{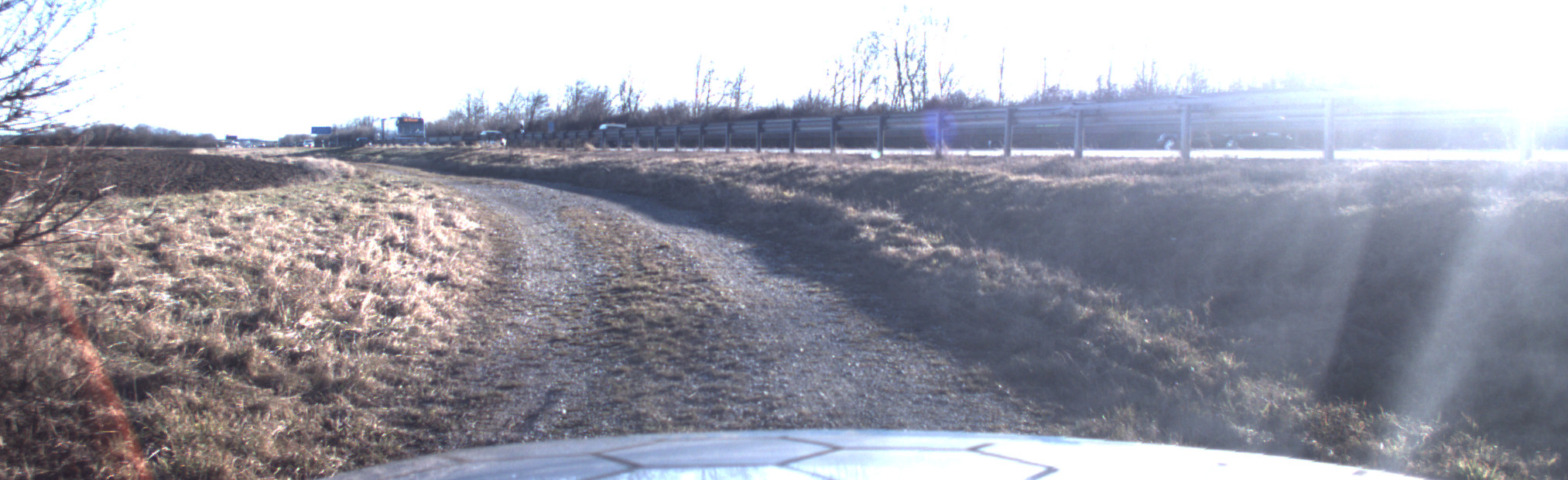}}}
            \fcolorbox{color24}{white}
            {\href{https://dataverse.nl/dataset.xhtml?persistentId=doi:10.34894/VIL0EV}{\includegraphics[height=\teaserimageheight]{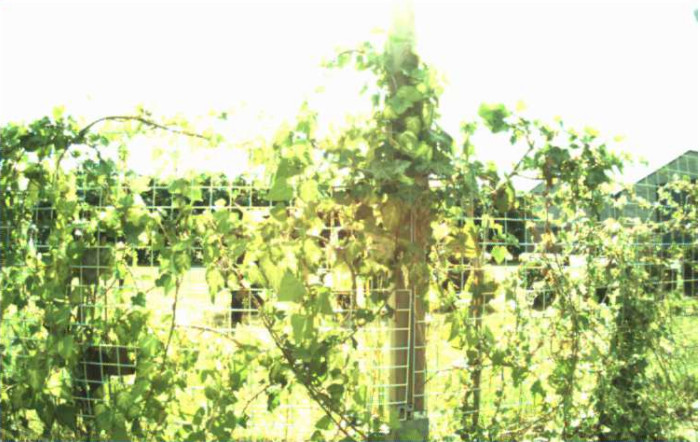}}}
            \fcolorbox{color25}{white}
            {\href{https://github.com/norlab-ulaval/logpiles_segmentation}{\includegraphics[height=\teaserimageheight]{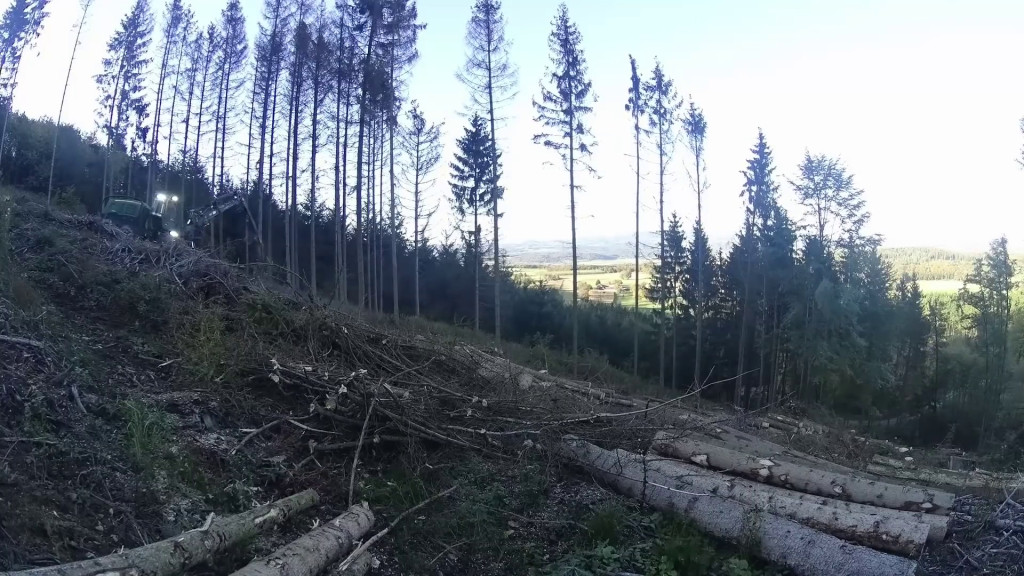}}}
            \fcolorbox{color26}{white}
            {\href{https://vgr.lab.yorku.ca/tools/trailnet/}{\includegraphics[height=\teaserimageheight]{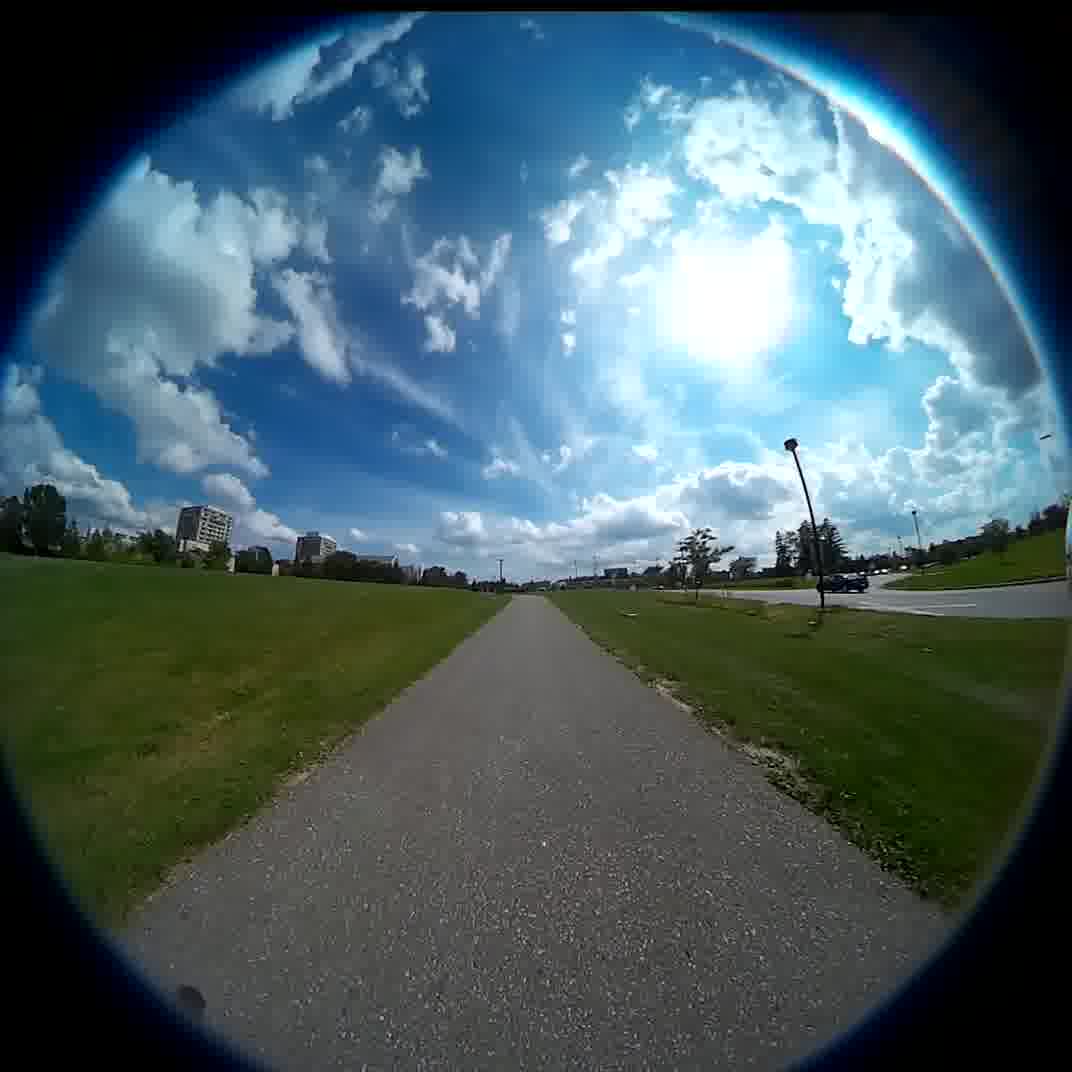}}}
            \fcolorbox{color27}{white}
            {\href{https://www.kaggle.com/datasets/sadhoss/vale-semantic-terrain-segmentation}{\includegraphics[height=\teaserimageheight]{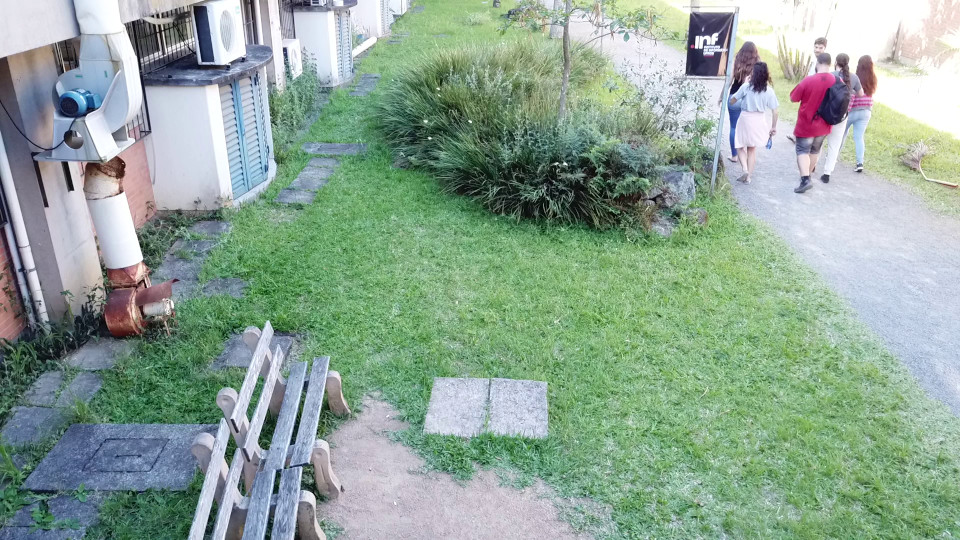}}}
            \fcolorbox{color28}{white}
            {\href{https://dataverse.orc.gmu.edu/dataset.xhtml?persistentId=doi:10.13021/orc2020/QSN50Q}{\includegraphics[height=\teaserimageheight]{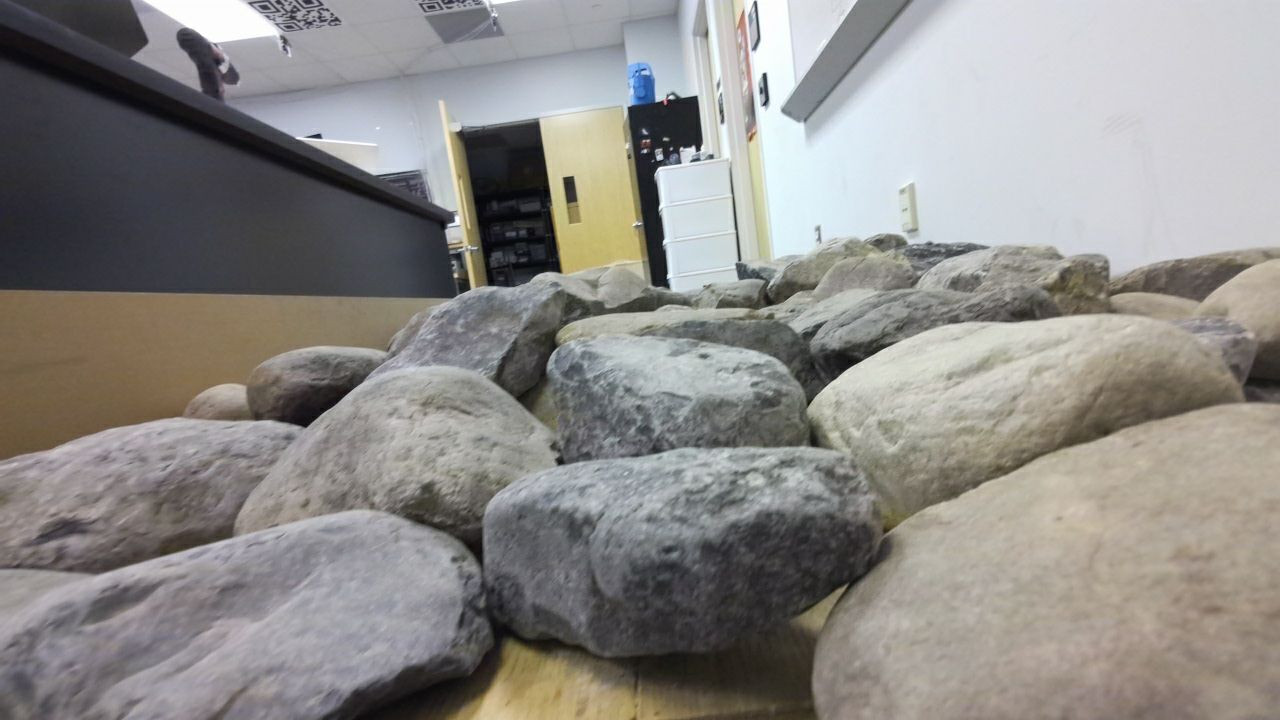}}}
            \fcolorbox{color29}{white}
            {\href{https://github.com/AerVisLoc/vpair}{\includegraphics[height=\teaserimageheight]{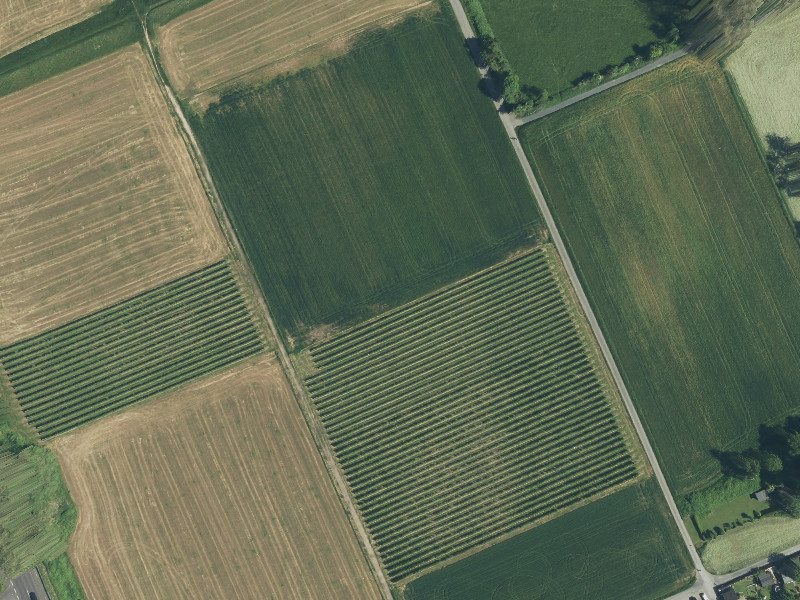}}}
		\fcolorbox{color30}{white}
            {\href{https://www.wilddash.cc/}{\includegraphics[height=\teaserimageheight]{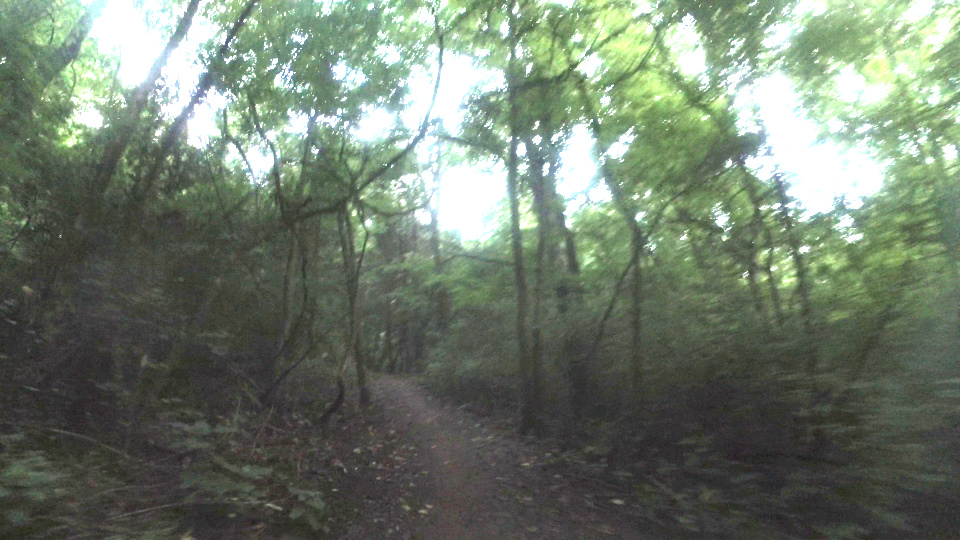}}}
		\fcolorbox{color31}{white}
            {\href{https://csiro-robotics.github.io/Wild-Places/}{\includegraphics[height=\teaserimageheight]{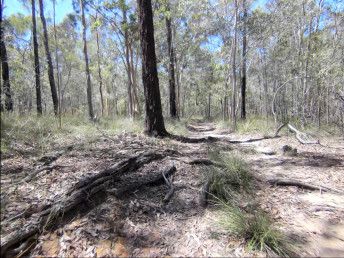}}}
		\fcolorbox{color32}{white}
            {\href{https://theairlab.org/yamaha-offroad-dataset/}{\includegraphics[height=\teaserimageheight]{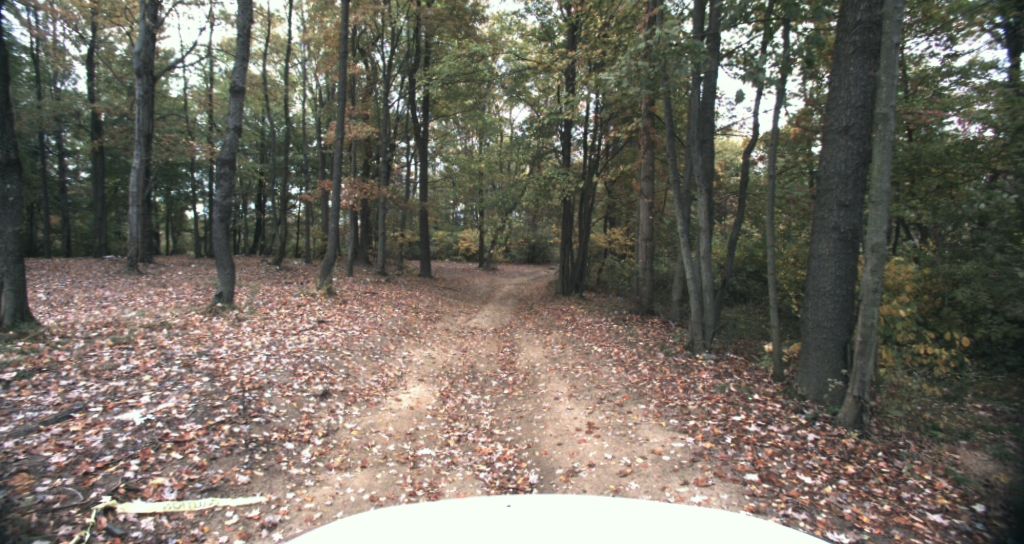}}}
			
			\scriptsize Fig. 1.~~An overview of the public research datasets of unstructured outdoor environments that are convered in this review: 
                \textcolor{color1}{\textbf{BarkNet 1.0}}~\cite{carpentier_barknet_2018}, 
			\textcolor{color2}{\textbf{BotanicGarden}}~\cite{liu_botanicgarden_2024}, 
			\textcolor{color3}{\textbf{CanaTree100}}~\cite{grondin_CanaTree100_2022}, 
			\textcolor{color4}{\textbf{CaSSeD}}~\cite{sharma_CaSSeD_2022},
			\textcolor{color5}{\textbf{CaT}}~\cite{sharma_CaT_2022},
			\textcolor{color6}{\textbf{EDEN}}~\cite{l2_eden_2021},
                \textcolor{color7}{\textbf{FinnWoodlands}}~\cite{lagos_finn_2023},
                \textcolor{color8}{\textbf{ForTrunkDet}}~\cite{dasilva_fortrunkdet_2021},
			\textcolor{color9}{\textbf{Freiburg Forest}}~\cite{valada_deepscene_2016},
			\textcolor{color10}{\textbf{GOOSE}}~\cite{mortimer_goose_2024},
                \textcolor{color11}{\textbf{GTASynth}}~\cite{curnis_gtasynth_2022},
			\textcolor{color12}{\textbf{MAVS}}~\cite{sharma_CaSSeD_2022},
			\textcolor{color13}{\textbf{Montmorency}}~\cite{tremblay_montmorency_2020},
			\textcolor{color14}{\textbf{OFFSED}}~\cite{neigel_offsed_2021},
			\textcolor{color15}{\textbf{ORFD}}~\cite{min_orfd_2022},
			\textcolor{color16}{\textbf{Rally Estonia}}~\cite{tampuu_estoniadriving_2023},
			\textcolor{color17}{\textbf{RELLIS-3D}}~\cite{jiang_rellis-3d_2021},
			\textcolor{color18}{\textbf{RUGD}}~\cite{wigness_rugd_2019},
                \textcolor{color19}{\textbf{SORT}}~\cite{yan_sort_2022},
                \textcolor{color20}{\textbf{SynPhoRest}}~\cite{nunes_synphorest_2022},
			\textcolor{color21}{\textbf{SynthTree43k}}~\cite{grondin_SynthTree43k_2022},
			\textcolor{color22}{\textbf{TartanDrive 2.0}}~\cite{sivaprakasam_tartandrive2_2024},
			\textcolor{color23}{\textbf{TAS500}}~\cite{metzger_tas500_2020},
			\textcolor{color24}{\textbf{TB-Places}}~\cite{leyvavallina_tbplaces_2019},
			\textcolor{color25}{\textbf{TimberSeg}}~\cite{fortin_timberseg_2022},
			\textcolor{color26}{\textbf{TrailNet}}~\cite{hoveidar_trailnet_2018},
			\textcolor{color27}{\textbf{Vale}}~\cite{hosseinpoor_vale_2021}.
			\textcolor{color28}{\textbf{Verti-Wheelers}}~\cite{datar_vertiwheelers_2024},
                \textcolor{color29}{\textbf{VPAIR}}~\cite{schleiss_vpair_2022},
			\textcolor{color30}{\textbf{Wilddash2}}~\cite{zendel_wilddash2_2022},
			\textcolor{color31}{\textbf{Wild-Places}}~\cite{knights_wildplaces_2023},
			\textcolor{color32}{\textbf{YCOR}}~\cite{maturana_ycor_2018}. \\
                Click on any image to open the download page for each of these datasets.
		\end{center}
	}]
}

\author{Peter Mortimer and Mirko Maehlisch\\
Institute for Autonomous Systems Technology\\
University of the Bundeswehr Munich\\
peter.mortimer@unibw.de}

\usepackage{fancyhdr}
\fancypagestyle{withfooter}{
  
  \fancyfoot[C]{\footnotesize Accepted to the IEEE ICRA Workshop on Field Robotics 2024}
}

\makeatletter
\let\NAT@parse\undefined
\makeatother
\usepackage[backref=page]{hyperref}
\hypersetup{colorlinks=true, 
	unicode=true, 
	linkcolor=[rgb]{0.10,0.05,0.67}, 
	citecolor=green, 
	filecolor=[rgb]{0.10,0.05,0.67}, 
	urlcolor=[RGB]{237, 15, 152} % url color copied from SemanticKITTI
}

\begin{document}
% Define the colors
\definecolor{color1}{RGB}{255, 0, 0}
\definecolor{color2}{RGB}{255, 38, 0}
\definecolor{color3}{RGB}{255, 75, 0}
\definecolor{color4}{RGB}{255, 113, 0}
\definecolor{color5}{RGB}{255, 150, 0}
\definecolor{color6}{RGB}{255, 188, 0}
\definecolor{color7}{RGB}{255, 225, 0}
\definecolor{color8}{RGB}{239, 247, 0}
\definecolor{color9}{RGB}{207, 254, 0}
\definecolor{color10}{RGB}{170, 255, 0}
\definecolor{color11}{RGB}{133, 255, 0}
\definecolor{color12}{RGB}{95, 255, 0}
\definecolor{color13}{RGB}{58, 255, 0}
\definecolor{color14}{RGB}{20, 255, 0}
\definecolor{color15}{RGB}{8, 255, 25}
\definecolor{color16}{RGB}{0, 255, 54}
\definecolor{color17}{RGB}{0, 255, 92}
\definecolor{color18}{RGB}{0, 255, 130}
\definecolor{color19}{RGB}{0, 255, 168}
\definecolor{color20}{RGB}{0, 255, 205}
\definecolor{color21}{RGB}{0, 245, 234}
\definecolor{color22}{RGB}{0, 226, 253}
\definecolor{color23}{RGB}{0, 191, 255}
\definecolor{color24}{RGB}{0, 154, 255}
\definecolor{color25}{RGB}{0, 116, 255}
\definecolor{color26}{RGB}{0, 78, 255}
\definecolor{color27}{RGB}{0, 40, 255}
\definecolor{color28}{RGB}{12, 15, 255}
\definecolor{color29}{RGB}{34, 0, 255}
\definecolor{color30}{RGB}{72, 0, 255}
\definecolor{color31}{RGB}{109, 0, 255}
\definecolor{color32}{RGB}{147, 0, 255}
\definecolor{color33}{RGB}{185, 0, 255}
\definecolor{color34}{RGB}{220, 0, 252}
\definecolor{color35}{RGB}{251, 0, 245}
\definecolor{color36}{RGB}{255, 0, 211}
\definecolor{color37}{RGB}{255, 0, 174}
\definecolor{color38}{RGB}{255, 0, 136}
\definecolor{color39}{RGB}{255, 0, 99}
\definecolor{color40}{RGB}{255, 0, 61}
\maketitle

\thispagestyle{withfooter}
\pagestyle{withfooter}

\begin{abstract}
Perception is an essential component of pipelines in field robotics.
In this survey, we quantitatively compare publicly available datasets available in unstructured outdoor environments.
We focus on datasets for common perception tasks in field robotics. 
Our survey categorizes and compares available research datasets.
This survey also reports on relevant dataset characteristics to help practitioners determine which dataset fits best for their own application.
We believe more consideration should be taken in choosing compatible annotation policies across the datasets in unstructured outdoor environments.
\end{abstract}

\IEEEpeerreviewmaketitle

\section{Introduction}

With the emergence of deep neural networks, the relevance of publicly available research data has increased. Public research datasets often serve two main purposes: for one they allow researchers to train their machine learning algorithms on the research datasets and demonstrate their application in novel scenarios.
This also gives smaller research groups access to learning algorithms, where the data collection and data annotation process is a high barrier to entry.
The second main purpose is as a benchmark to evaluate different perception tasks, allowing for a level of quantitative comparison and reproducibility in Computer Vision that was previously unavailable. 
Commonly, this comparison is done by withholding a test split of the dataset that is used to evaluate the predictive performance of a method. 

%For autonomous driving in urban environments, public research datasets like KITTI~\cite{geiger_vision_2013}, CityScapes~\cite{cordts_cityscapes_2016} and NuScenes~\cite{caesar_nuscenes_2020} have established themselves as popular training and evaluation datasets. 
For unstructured outdoor environments research datasets exist, but unlike to their urban counterparts, none of the datasets provide a public leaderboard or evaluation server.
This is for one due to the smaller research community that focuses on perception in unstructured outdoor environments, but we also think the diversity of applications that operate in unstructured outdoor environments require different datasets and a higher diversity of datasets.

This motivated us to capture a snapshot of the available research datasets recorded in unstructured outdoor environments in this survey.
Recent surveys~\cite{guastella_learning_methods_ugv_review_2021,islam_offroad_detection_review_2022, wijayathunga_challenges_ugv_2023} laid a greater focus on organizing all components of the typical Sense-Plan-Act framework involved to operate autonomous ground vehicles in unstructured environments.
We decided to narrow our focus on the available research datasets and providing comparisons that should help researchers decide on which datasets to use for their application.
This is best done by quantitatively comparing relevant dataset characteristics.

%This paper makes the following contributions:
%\begin{itemize}
%    \item We provide a comprehensive overview of publicly available research datasets in unstructured outdoor environments.
%    \item We quantitatively compare relevant dataset characteristics about the sensor data and the robot platform.
    % \item We provide a prospective view of the potential of foundation models like SEEM~\cite{zou_seem_2023} in enhancing current research datasets with autolabeled data for supervised learning algorithms.
%\end{itemize}

\begin{table*}[htpb!]
\caption*{Comparison of 2D Semantic Segmentation Datasets}
\begin{tabular}{@{}lccccccc@{}}
\toprule
Dataset & Robot Platform & Image Modalities & \# Images & \# Classes & $\text{z}_{\text{ground}}$ & FOV$_{\text{h}}$ & FOV$_{\text{v}}$ \\ \midrule
OFFSED \cite{neigel_offsed_2021} & ZED & RGB, stereo depth & 203 & 19 & - & $76.45^\circ$ & $47.79^\circ$ \\
Freiburg Forest \cite{valada_deepscene_2016} & VIONA & RGB, multi-spectral, stereo depth & 366 & 7 & \SI{1.66}{\meter} & $66^\circ$ & - \\
% This is assuming that the BumbleBee left camera was used and that the frame base_footprint resembles 
% the base minus it's height over a flat surface (e.g. rosrun tf tf_echo bumblebee1 base_footprint )
% the ROS Bag was taken from the LifeNav project done at Freiburg university.
% The FoV is taken from the Bumblebee2 data sheet for the 38mm focal length of the BB2-08S2 model.
% This unfortunately does not consider, that the image is cropped and will be corrected in the final version.
TAS500 \cite{metzger_tas500_2020} & MuCAR-3 & RGB & 640 & 24 & \SI{1.52}{\meter} & $46.28^\circ$ & $14.74^\circ$ \\
YCOR \cite{maturana_ycor_2018} & ATV & RGB & 1\,076 & 8 & \SI{1.94}{\meter} & $40.13^\circ$ & $21.96^\circ$ \\
% the height is based on the specifications of the 2015 Viking VI ATV
% https://web.yamahamotorsports.com/assets/pages/2015_Yamaha_Viking_VI_Brochure.pdf
% the master thesis of John Mai (CMU) mentions using the "2014 model Yamaha Viking VI side-by-side UTV"
% The TartanDrive2 paper mentions this vehicle being replaced by a newer model
BoatanicGarden \cite{liu_botanicgarden_2024} & Scout V1.0 & RGB, stereo depth & 1181 & 27 & \SI{0.98}{\meter} & $71^\circ$ & $56^\circ$ \\
% extrinsic calibration found here: https://github.com/robot-pesg/BotanicGarden/blob/main/calib/extrinsics/calib_chain.yaml
CaSSeD \cite{sharma_CaSSeD_2022} & ATV & RGB & 1\,679 & 6 & \SI{2.10}{\meter} & $60^\circ$ & $39.77^\circ$ \\
MAVS \cite{sharma_CaSSeD_2022} & simulation & RGB & 1\,974 & 6 & - & - & - \\ 
SynPhoRest~\cite{nunes_synphorest_2022} & simulation & RGB, Depth, LiDAR & 3\,154 & 6 & - & - & - \\
CaT \cite{sharma_CaT_2022} & ATV & RGB & 3\,624 & 3 & \SI{2.10}{\meter} & $60^\circ$ & $39.77^\circ$ \\
RELLIS-3D \cite{jiang_rellis-3d_2021} & Warthog & RGB, stereo depth & 6\,235 & 20 & \SI{0.97}{\meter} & $37.68^\circ$ & $24.12^\circ$ \\
RUGD \cite{wigness_rugd_2019} & Husky & RGB  & 7\,546 & 24 & \SI{0.25}{\meter} & $42.66^\circ$  & $34.96^\circ$ \\
% get the data for the camera from the paper, there does not seem to be a ROS Bag file or the camera intrinsics published
% the paper mentions "camera sensor is mounted on the front of the platform just above the external height of the robot, 
% resulting in an nvironment viewpoint from less than 25 centimeters off the ground"
GOOSE \cite{mortimer_goose_2024} & MuCAR-3 & RGB, NIR & 8\,790 & 64 & \SI{1.57}{\meter} & $46.56^\circ$ & $31.30^\circ$ \\ \bottomrule
\end{tabular}
\centering
\caption{
The datasets are sorted in increasing size of annotated images.
The number of annotated images only accounts for the data available to the user. 
The test split that is often hidden for evaluation purposes is therefore excluded.
For the $\text{z}_{\text{ground}}$, FOV$_{\text{h}}$, FOV$_{\text{v}}$ some entries are missing when the extrinsic and intrinsic camera information was missing. 
For the final version, these entries can be estimated.
}
\label{tab:2d_segmentation_datasets}
\end{table*}

\newlength{\semsegimagesize}
\setlength{\semsegimagesize}{3.75em}
\begin{figure*}[htpb!]
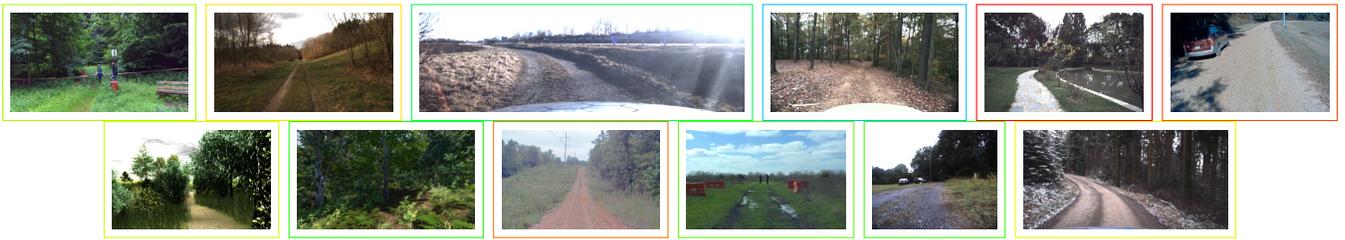

\begin{center}
   \rule[1ex]{.9\linewidth}{0.5pt} \\
\end{center}

\begin{center}
\fcolorbox{color10}{white}
{\href{https://www.dfki.uni-kl.de/~neigel/offsed.html}{\includegraphics[height=\semsegimagesize]{images/offsed_left_184.jpg}}} 
\fcolorbox{color7}{white}
{\href{http://deepscene.cs.uni-freiburg.de\#datasets}{\includegraphics[height=\semsegimagesize]{images/freiburg_forest_b85-1576_Clipped.jpg}}}
\fcolorbox{color17}{white}
{\href{https://mucar3.de/icpr2020-tas500/}{\includegraphics[height=\semsegimagesize]{images/TAS500_1580998311720168192.jpg}}}
\fcolorbox{color23}{white}
{\href{https://theairlab.org/yamaha-offroad-dataset/}{\includegraphics[height=\semsegimagesize]{images/YCOR_train_iid000022.jpg}}}
\fcolorbox{color1}{white}
{\href{https://github.com/ulaval-damas/tree-bark-classification?tab=readme-ov-file\#barknet-10-database}{\includegraphics[height=\semsegimagesize]{images/BotanicGarden_c_01.jpg}}}
\fcolorbox{color3}{white}
{\href{https://www.cavs.msstate.edu/resources/autonomous_dataset.php}{\includegraphics[height=\semsegimagesize]{images/CaSSed_Dataset3_1570151386.067731344.jpg}}}
\fcolorbox{color9}{white}
{\href{https://www.cavs.msstate.edu/resources/autonomous_dataset.php}{\includegraphics[height=\semsegimagesize]{images/MAVS_image_1001.jpg}}}
\fcolorbox{color14}{white}{\href{https://zenodo.org/records/6369446}
{\includegraphics[height=\semsegimagesize]{images/SynPhoRest.jpg}}}
\fcolorbox{color4}{white}
{\href{https://www.cavs.msstate.edu/resources/autonomous_dataset.php}{\includegraphics[height=\semsegimagesize]{images/CaT_Power_Line_Train_img_402.jpg}}}
\fcolorbox{color12}{white}
{\href{https://github.com/unmannedlab/RELLIS-3D}{\includegraphics[height=\semsegimagesize]
{images/Rellis-3D_00003_frame000118-1581624087_050.jpg}}}
\fcolorbox{color13}{white}{\href{http://rugd.vision/}
{\includegraphics[height=\semsegimagesize]{images/RUGD_trail-11_01361.jpg}}}
\fcolorbox{color8}{white}
{\href{https://goose-dataset.de/}{
\includegraphics[height=\semsegimagesize]{images/goose_image_gallery.jpg}}}
\caption*{
An example image of each dataset presented in the same order as in Table~\ref{tab:2d_segmentation_datasets}.
Click on any image to open the download page for each of these datasets. Listed are
\textcolor{color10}{\textbf{OFFSED}}~\cite{neigel_offsed_2021},
\textcolor{color7}{\textbf{Freiburg Forest}}~\cite{valada_deepscene_2016},
\textcolor{color17}{\textbf{TAS500}}~\cite{metzger_tas500_2020},
\textcolor{color23}{\textbf{YCOR}}~\cite{maturana_ycor_2018},
\textcolor{color2}{\textbf{BotanicGarden}}~\cite{liu_botanicgarden_2024}, 
\textcolor{color3}{\textbf{CaSSeD}}~\cite{sharma_CaSSeD_2022},
\textcolor{color9}{\textbf{MAVS}}~\cite{sharma_CaSSeD_2022},
    \textcolor{color14}{\textbf{SynPhoRest}}~\cite{nunes_synphorest_2022},
\textcolor{color4}{\textbf{CaT}}~\cite{sharma_CaT_2022},
\textcolor{color12}{\textbf{RELLIS-3D}}~\cite{jiang_rellis-3d_2021},
\textcolor{color13}{\textbf{RUGD}}~\cite{wigness_rugd_2019},
\textcolor{color8}{\textbf{GOOSE}}~\cite{mortimer_goose_2024}.
}

\end{center}

\begin{center}
    \rule[1ex]{.9\linewidth}{0.5pt} \\
\end{center}

\end{figure*}

\section{Datasets}

\subsection{2D Semantic Segmentation}

Image datasets with pixel-wise annotations of their surrounding make up the largest group of publicly available research datasets in unstructured outdoor environments.
In Table~\ref{tab:2d_segmentation_datasets}, we organize all relevant datasets to the best of our knowledge. 
Systematic analysis have shown the inductive biases learned by machine learning algorithms~\cite{dijk_mde_analysis_2019}.
These biases can be traced back to dataset characteristics like the position and orientation of the camera on the robotic platform.
To account for this, we also additionally provide the following dataset characteristics in Table~\ref{tab:2d_segmentation_datasets}:\\
\noindent\textbf{height above ground ($\text{z}_{\text{ground}}$):} the $\text{z}_{\text{ground}}$ is the height of the mounted sensor camera above the ground surface that the robot platform is traversing on. 
This determines the height of the horizon line in the camera image. \\
\noindent\textbf{horizontal field of view (FOV$_{\text{h}}$):} the FOV$_{\text{h}}$ describes the extent of the observable world along the camera's horizontal axis in degrees. 
The FOV$_{\text{h}}$ can be calculated using the focal length along the horizontal axis $f_x$ and the width of the image sensor $w$ in pixels:
\begin{equation*}
    \text{FOV}_{\text{h}} = 2 \cdot \arctan\left(\frac{w}{2f_x}\right)
\end{equation*}
\\
\noindent\textbf{vertical field of view (FOV$_{\text{v}}$):} the FOV$_{\text{v}}$ describes the extent of the observable world along the camera's vertical axis in degrees. 
The FOV$_{\text{v}}$ can be calculated using the focal length along the vertical axis $f_y$ and the height of the image sensor $h$ in pixels: 
\begin{equation*}
    \text{FOV}_{\text{v}} = 2 \cdot \arctan\left(\frac{h}{2f_y}\right)
\end{equation*}
\\
These calculations approximate the optical system using a pinhole camera model.
An increase in field of view for most optical systems on robotic platforms also entails distortion effects the farther away the pixel is from the optical axis. 
Also effects in the appearance of colors can differ due to spherical aberration.
Taking these effects into account can be relevant when the perception pipeline relies on neural networks pretrained on images with a different camera setup.

A few general trends can be observed from the 2D semantic segmentation datasets listed in Table~\ref{tab:2d_segmentation_datasets}.
Over time, the size and the annotation granularity is increasing.
The first datasets in this domain in the likes of OFFSED~\cite{neigel_offsed_2021} and Freiburg Forest~\cite{valada_deepscene_2016} have less than 1\,000 annotated images and Freiburg Forest uses less than 10 semantic classes to describe the scenes in the dataset.

\begin{table*}[htpb!]
\caption*{Comparison of 3D LiDAR Semantic Segmentation Datasets}
\begin{tabular}{@{}lcccccccc@{}}
\toprule
Dataset & Robot Platform & LiDAR Scanner & \# Channels & \# Annotations & \# Classes & $\text{z}_{\text{ground}}$ & FOV$_{\text{h}}$ & FOV$_{\text{v}}$ \\ \midrule

SynPhoRest~\cite{nunes_synphorest_2022} & simulation & Livox Mid-40 & non-repetitive circular & 3\,154 & 6 & - & $38.4^\circ$ & $38.4^\circ$ \\

GOOSE \cite{mortimer_goose_2024} & MuCAR-3 & Velodyne Alpha Prime & 128 & 8\,790 & 64 & \SI{2.23}{\meter} & $360^\circ$ & $40^\circ$ \\ 
% vertical FoV is based on the Velodyne Alpha Prime Datasheet (https://www.goetting.de/dateien/downloads/VelodyneLidar_AlphaPrime_Datasheet.pdf)

RELLIS-3D \cite{jiang_rellis-3d_2021} & Warthog & Ouster OS1 & 64 & 13\,556 & 20 & \SI{1.13}{\meter} & $360^\circ$ & $45^\circ$ \\
% vertical FoV is based on the Ouster OS1 Datasheet (https://data.ouster.io/downloads/datasheets/datasheet-revd-v2p0-os1.pdf)
\bottomrule
\end{tabular}
\centering
\caption{
All datasets were recorded with a rotating LiDAR scanner mounted on the top of the robot platform, allowing for a $360^\circ$ view of the environment around the robot platform with little self-occlusion.
The LiDAR scanner used in the GOOSE dataset has 128 channels per firing, leading to a more dense point cloud that makes objects in the far distance more easily visible than the LiDAR scanner used in RELLIS-3D.
The FOV$_{\text{v}}$ was determined for each dataset by reviewing the datasheet for the particular LiDAR scanner. 
}
\label{tab:3d_segmentation_datasets}
\end{table*}

\newlength{\semsegpcdsize}
\setlength{\semsegpcdsize}{7em}
\begin{figure*}[htpb!]
\begin{center}
   \rule[1ex]{.9\linewidth}{0.5pt} \\
\end{center}

\begin{center}
\fcolorbox{color14}{white}
{\includegraphics[height=\semsegpcdsize]{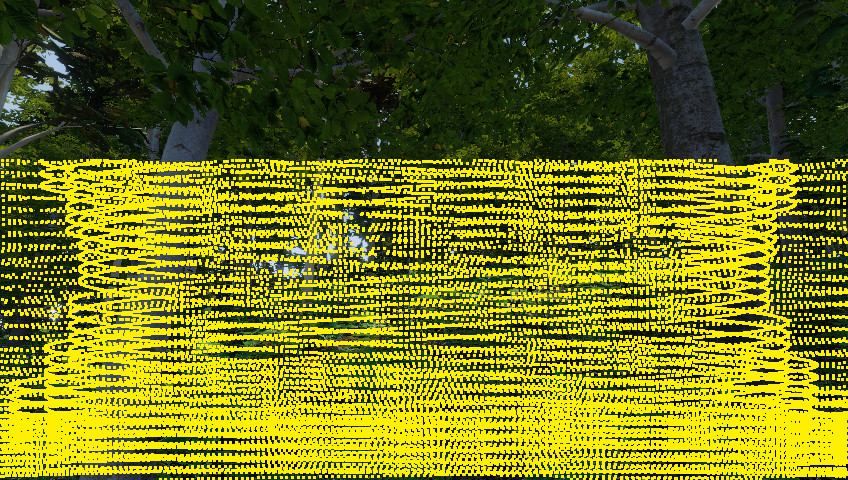}} 
\fcolorbox{color12}{white}
{\includegraphics[height=\semsegpcdsize]{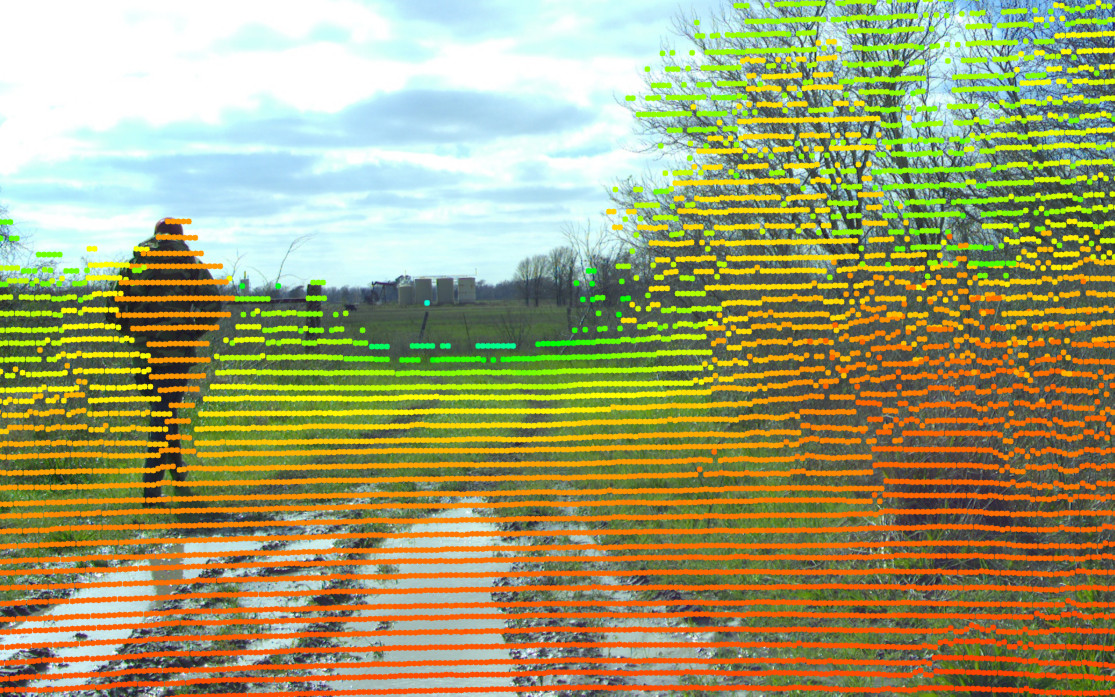}} 
\fcolorbox{color8}{white}
{\includegraphics[height=\semsegpcdsize]{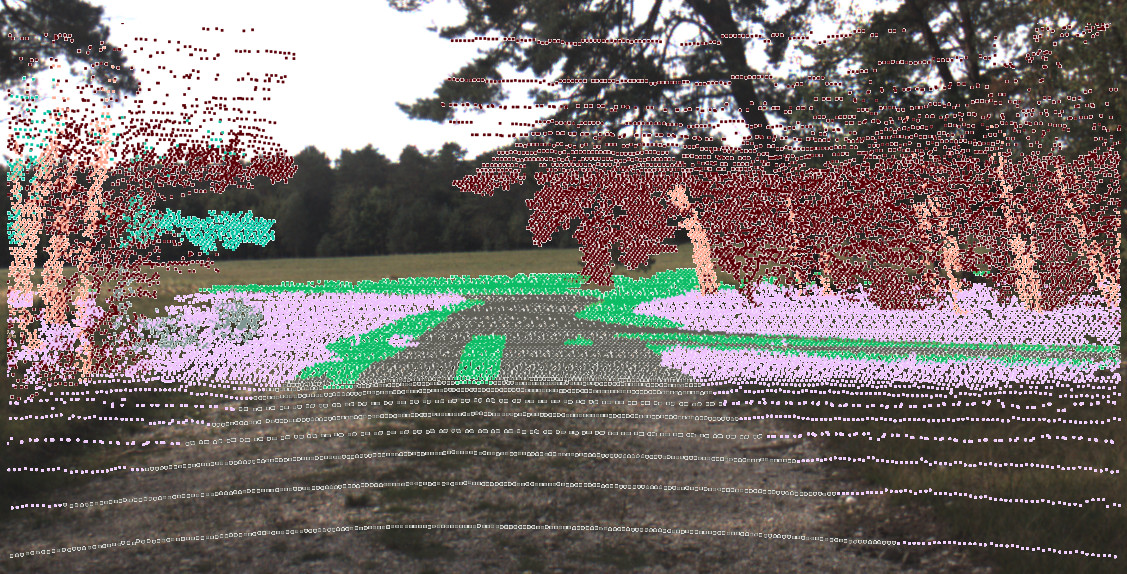}}
\end{center}

\caption*{
Images with the LiDAR points projected into the scenes of \textcolor{color14}{\textbf{SynPhoRest}}~\cite{nunes_synphorest_2022}, \textcolor{color12}{\textbf{RELLIS-3D}}~\cite{jiang_rellis-3d_2021} and \textcolor{color8}{\textbf{GOOSE}}~\cite{mortimer_goose_2024}.
The LiDAR in SynPhoRest is modeled after a LiDAR scanner with a non-repetitive scanning pattern, leading to a different pattern in comparison to the rotating LiDAR scanner observed in RELLIS-3D and GOOSE. 
}

\begin{center}
    \rule[1ex]{.9\linewidth}{0.5pt} \\
\end{center}

\end{figure*}

In comparison, more recent datasets like CaSSeD~\cite{sharma_CaSSeD_2022}, RELLIS-3D~\cite{jiang_rellis-3d_2021}, RUGD~\cite{wigness_rugd_2019} and GOOSE~\cite{mortimer_goose_2024} have sample sizes just below 10\,000 images and often over 20 different semantic classes.
This trend of increasing the sample size and semantic granularity also increases the annotation effort, often completed by human annotators.
Especially for fully annotated semantic masks, the average annotation time can exceed 1 hour for a human annotator~\cite{cordts_cityscapes_2016}. \\
There are some exceptions to this trend of increasing the annotation effort with each released outdoor dataset. For instance, CaT~\cite{sharma_CaT_2022} annotates the images based on the traversability assessment for the three vehicle types $\{$\textit{Sedan}, \textit{Pickup},  \textit{Off-road}$\}$.
This differs from the common practice in semantic segmentation datasets to segment the scene into stuff and thing classes based on their direct visual appearance~\cite{caesar_cocostuff_2018}.
In the context of field robotics, the detection of trafficable ground surfaces is a common task and the annotation policy of CaT directly encodes this trafficability information for different vehicle types.
This avoids the intermediate step of first segmenting the scene into different ground surface types like $\{$\textit{grass}, \textit{gravel},  \textit{soil}, ...$\}$, which have to be correctly inferred for the particular robot platform considering properties like ground clearance and wheel size.
This approach to annotate the scenes of a dataset closer to the task at hand can be extended to other perception tasks like segmenting moving objects~\cite{vertens_smsnet_2017, chen_mos_2022} and their intentions.
In subsection~\ref{subsec:terrain_datasets}, we compare existing datasets intended for segmenting the terrain surface.
% add line about the difference in z_ground between the GOOSE and RELLIS-3D.
% In general, there are very few LiDAR semantic segmentation datasets available for unstructured outdoor environments.

\begin{table*}[htpb!]
\caption*{Comparison of Terrain Datasets}
\begin{tabular}{@{}lcccc@{}}
\toprule
Dataset & Vehicle & Sensors & GT Surfaces & Surface Types \\ \midrule
TrailNet~\cite{hoveidar_trailnet_2018} & Pioneer 3-AT & Camera & \xmark & \textit{asphalt}, \textit{cobblestone}, \textit{concrete}, \textit{dirt}, \textit{gravel} \\
ORFD~\cite{min_orfd_2022} & Mazda Ruiyi 6 Coupe & Camera, LiDAR & \cmark & \textit{Countryside}, \textit{Farmland}, \textit{Grassland}, \textit{Woodland} \\ 
Rally Estonia~\cite{tampuu_estoniadriving_2023} &  Lexus RX & Camera, LiDAR & \xmark & \textit{gravel roads}, \textit{paved roads} \\ 
Vale~\cite{hosseinpoor_vale_2021} & DJI Mavic Pro drone & Camera & \cmark & \textit{Wheeled}, \textit{Tracked}, \textit{Legged}, \textit{Non-Traversable} \\ 
CaT~\cite{sharma_CaT_2022} & Polaris Ranger Crew XP 100 & Camera & \cmark & \textit{Sedan}, \textit{Pickup}, \textit{Off-road} \\
Verti-Wheelers~\cite{datar_vertiwheelers_2024} & V4W \& V6W & Kinect Camera & \xmark & \textit{indoor testbed}, \textit{rocky outdoor environment} \\
\bottomrule
\end{tabular}
\centering
\caption{
Comparison of terrain datasets that focus on the detection of offroad paths or the trafficability surfaces in unstructured outdoor environments. 
The groundtruth surfaces in the ORFD dataset resemble the free space on the road, where the vehicle can operate on while avoiding obstacles on the road. 
In the case of the Vale and CaT datasets, the groundtruth annotations describe the trafficability of the surrounding terrain for different types of robot locomotion or vehicle types with increasing ride heights. 
}
\label{tab:terrain_datasets}
\end{table*}

\newlength{\terrainimagesize}
\setlength{\terrainimagesize}{4.2em}
\begin{figure*}[htpb!]
\begin{center}
   \rule[1ex]{.9\linewidth}{0.5pt} \\
\end{center}

\begin{center}
% TrailNet
\fcolorbox{color19}{white}
{\href{https://vgr.lab.yorku.ca/tools/trailnet/}{\includegraphics[height=\terrainimagesize]{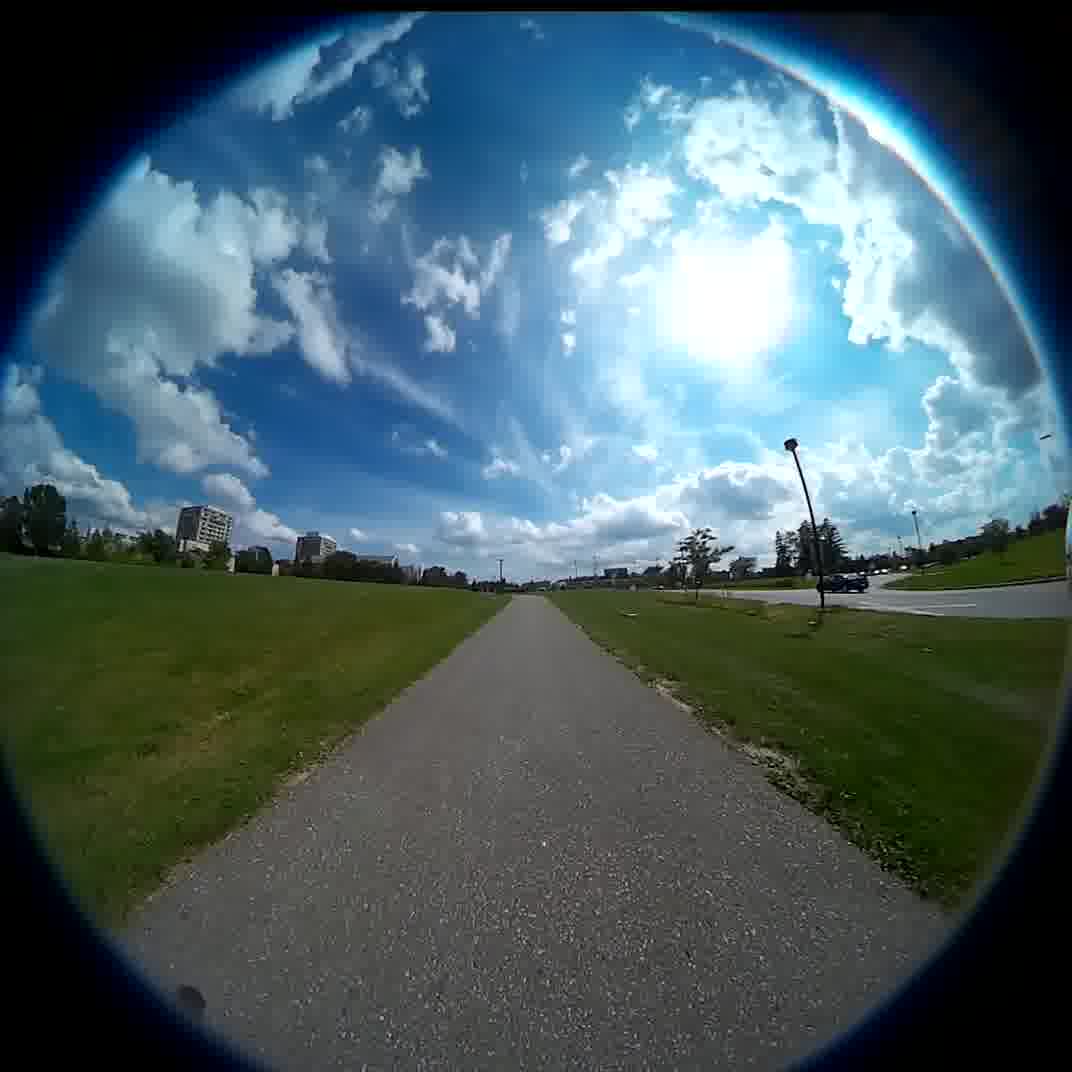}}}
% ORFD
\fcolorbox{color11}{white}
{\href{https://github.com/chaytonmin/Off-Road-Freespace-Detection\#Prepare-data}{\includegraphics[height=\terrainimagesize]{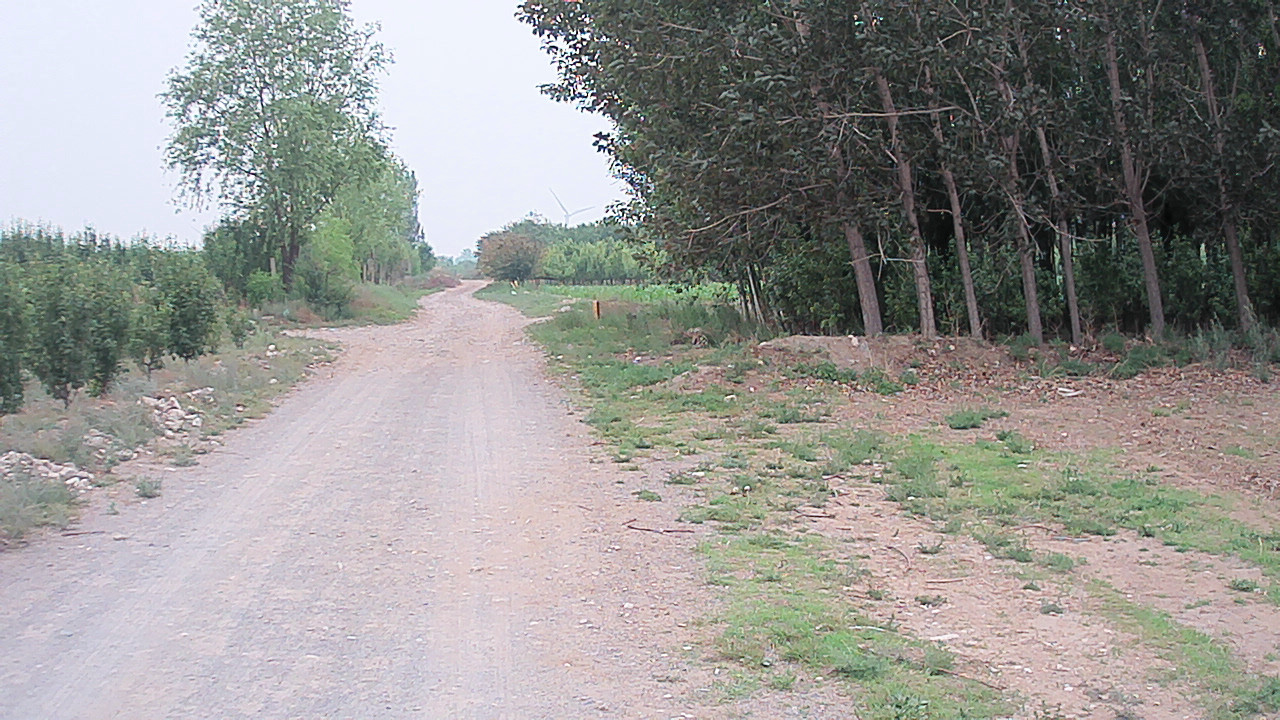}}}
% Rally Estonia
\fcolorbox{color5}{white}
{\href{https://github.com/UT-ADL/e2e-rally-estonia\#dataset}{\includegraphics[height=\terrainimagesize]{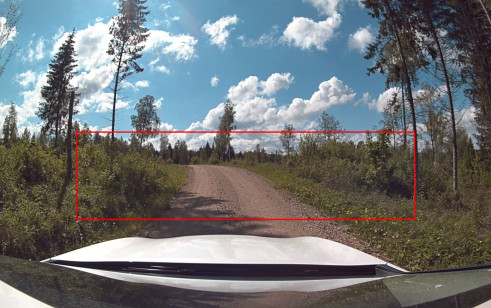}}}
% Vale
\fcolorbox{color20}{white}
{\href{https://www.kaggle.com/datasets/sadhoss/vale-semantic-terrain-segmentation}{\includegraphics[height=\terrainimagesize]{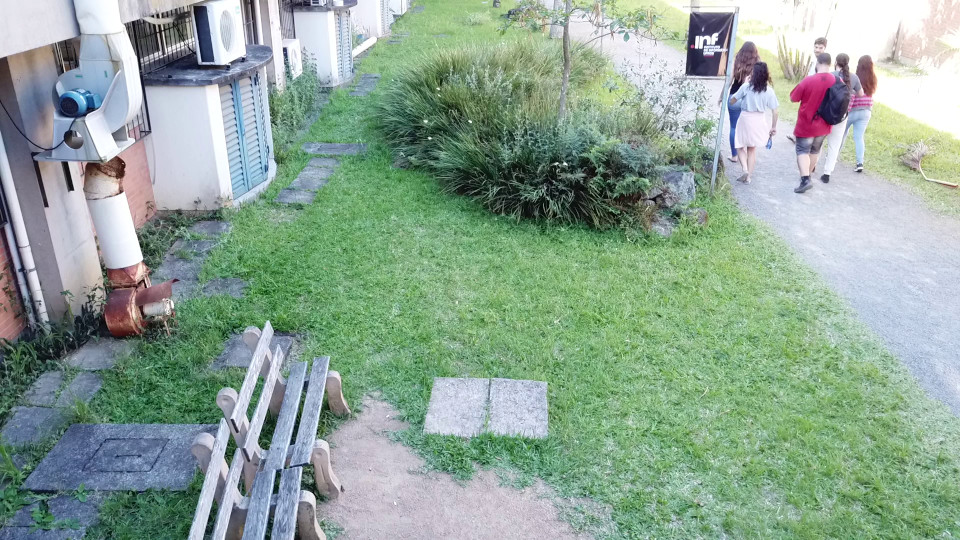}}}
% CaT
\fcolorbox{color4}{white}
{\href{https://www.cavs.msstate.edu/resources/autonomous_dataset.php}{\includegraphics[height=\terrainimagesize]{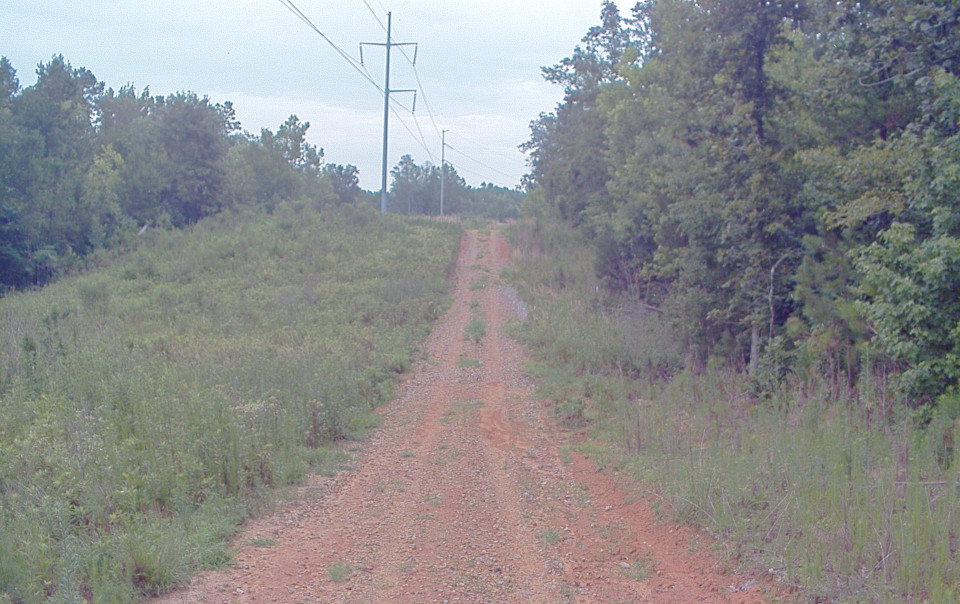}}}
% Verti-Wheelers
\fcolorbox{color20}{white}
{\href{https://dataverse.orc.gmu.edu/dataset.xhtml?persistentId=doi:10.13021/orc2020/QSN50Q}{\includegraphics[height=\terrainimagesize]{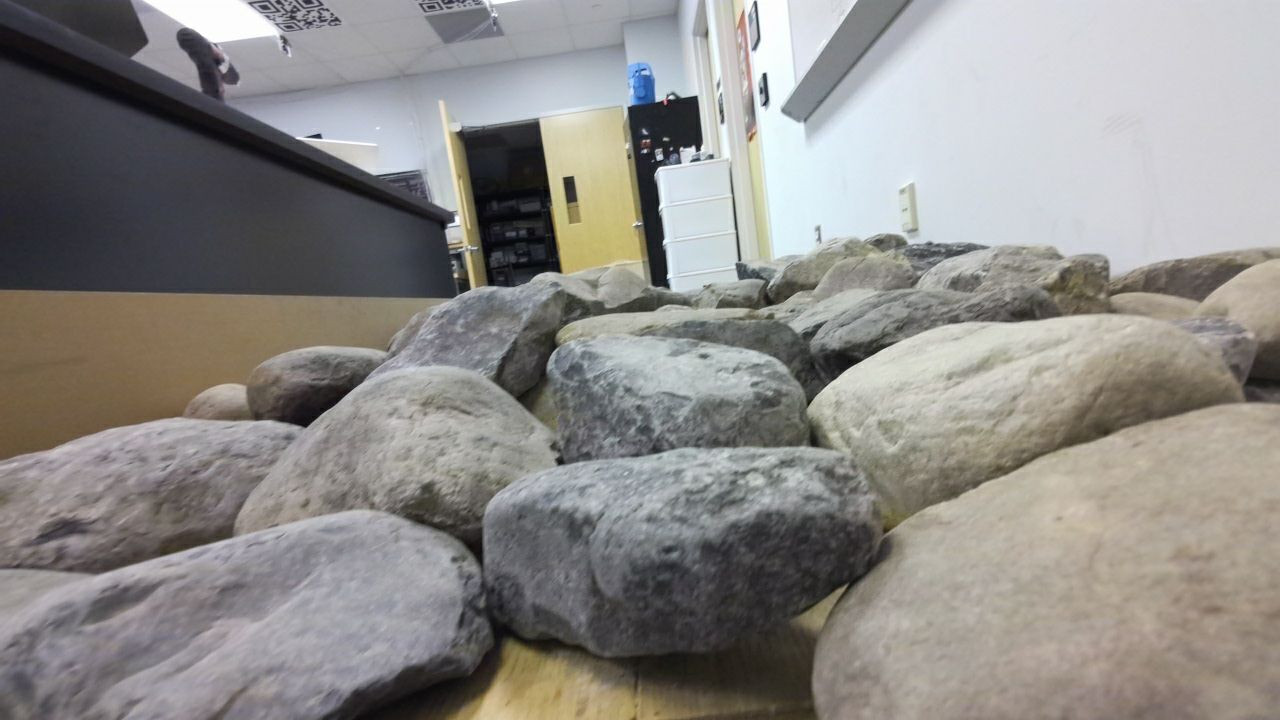}}}

\caption*{
An example image of each dataset presented in the same order as in Table~\ref{tab:terrain_datasets}.
Click on any image to open the download page for each of these datasets. Listed are
\textcolor{color19}{\textbf{TrailNet}}~\cite{hoveidar_trailnet_2018},
\textcolor{color11}{\textbf{ORFD}}~\cite{min_orfd_2022},
\textcolor{color5}{\textbf{Rally Estonia}}~\cite{tampuu_estoniadriving_2023},
\textcolor{color23}{\textbf{Vale}}~\cite{hosseinpoor_vale_2021}.
\textcolor{color4}{\textbf{CaT}}~\cite{sharma_CaT_2022},
\textcolor{color20}{\textbf{Verti-Wheelers}}~\cite{datar_vertiwheelers_2024}.
}

\end{center}

\begin{center}
    \rule[1ex]{.9\linewidth}{0.5pt} \\
\end{center}

\end{figure*}

\subsection{3D LiDAR Semantic Segmentation}

Additional to camera images, the point clouds generated from LiDAR scans have over time become another common scene representation that is semantically segmented for learning algorithms.
For urban environments, datasets like Semantic3D~\cite{hackel_semantic3d_2017}, SemanticKITTI~\cite{behley_semantickitti_2019} and NuScenes~\cite{caesar_nuscenes_2020} were released. 
Table~\ref{tab:3d_segmentation_datasets} gives an overview of all semantic segmentation datasets recorded in unstructured outdoor environments. 
Generally, the 3D point cloud annotation requires more effort by human annotators as reference images and 3D point cloud annotation tools have to be used to correctly annotate the LiDAR point cloud data. \\
RELLIS-3D~\cite{jiang_rellis-3d_2021} is a large dataset consisting of over 10\,000 annotated LiDAR scans recorded in five sequences at a ground research facility on the Rellis Campus of Texas A\&M University.
The LiDAR point cloud in RELLIS-3D was refined by annotating subsequent LiDAR scans from the same scene.
The initial annotation of the point cloud was generated by projecting the annotated pixels of the camera images onto the nearest point in the point cloud. 
The 3D point cloud annotation tool by SemanticKITTI~\cite{behley_semantickitti_2019} is openly available to annotate aggregated LiDAR scans.
This annotation approach requires a good camera-LiDAR calibration, the correct alignment of subsequent LiDAR scans using a highly accurate INS or a SLAM system to loop close sequences and a process to handle dynamic objects that are detected over multiple LiDAR scans.
When these conditions are met, a large set of LiDAR scans can be annotated efficiently, leading to the large number of annotated scans available for RELLIS-3D. \\
The GOOSE dataset~\cite{mortimer_goose_2024} contains annotated LiDAR scans across 234 recorded sequences.
Within a sequence, LiDAR scans are annotated after every \SIrange[range-phrase={-}]{5}{10}{\second} of driving.
The sequences are recorded over the course of the year, leading to a more diverse set of weather conditions and outdoor environments.
The annotation effort to annotate effectively separate scenes for each LiDAR scan is higher than the annotation approach used for RELLIS-3D.

\subsection{Synthetic Datasets}

\begin{table}[htpb!]
\caption*{Comparison of Synthetic Outdoor Datasets}
\begin{tabular}{@{}lcccc@{}}
\toprule
Dataset & Environments & Camera & LiDAR & \# Scenes \\
\midrule 
\multirow{2}{*}{MVAS \cite{sharma_CaSSeD_2022}} & \multirow{2}{*}{Forest, Grassland} & \multirow{2}{*}{\cmark} & \multirow{2}{*}{\xmark} & \multirow{2}{*}{1\,974} \\
& & & & \\
\multirow{2}{*}{SynPhoRest \cite{nunes_synphorest_2022}} & \multirow{2}{*}{Forest} & \multirow{2}{*}{\cmark} & \multirow{2}{*}{\cmark} & \multirow{2}{*}{3\,154} \\
& & & & \\
\multirow{2}{*}{GTASynth \cite{curnis_gtasynth_2022}} & \multirow{2}{*}{\parbox{2.4cm}{\centering Canyon, Field, Hill \\ Park, River}} & \multirow{2}{*}{\cmark} & \multirow{2}{*}{\cmark} & \multirow{2}{*}{16\,207} \\
& & & & \\
\multirow{2}{*}{SORT \cite{yan_sort_2022}} & \multirow{2}{*}{\parbox{2.4cm}{\centering Desert, Forest, \\ Grassland, Mountain}} & \multirow{2}{*}{\cmark} & \multirow{2}{*}{\xmark} & \multirow{2}{*}{20\,844} \\
& & & & \\
\multirow{2}{*}{EDEN \cite{l2_eden_2021}} & \multirow{2}{*}{Garden} & \multirow{2}{*}{\cmark} & \multirow{2}{*}{\xmark} & \multirow{2}{*}{369\,663} \\
& & & & \\
\bottomrule
\end{tabular}
\centering
\caption{
MVAS, SynPhoRest, SORT are semantic segmentation datasets created in virtual outdoor environments.
They provide pixel-perfect semantic masks of their image data and SynPhoRest even simulates the firings of a LiDAR scanner for semantically segmented 3D point cloud data.
GTASynth is a registration and SLAM dataset that can provide a ground truth position and orientation.
}
\label{tab:synthetic_datasets}
\end{table}
\vspace{1em}
Synthetic outdoor datasets like MVAS~\cite{sharma_CaSSeD_2022} makes use of a simulation environment, where the geometries in the scene can be perfectly traced back to the semantic class they belong to.
The images from MVAS are not photorealistic (see the figure under Table~\ref{tab:2d_segmentation_datasets}).
% Fix that the figure number is not visible
Simulators for autonmous driving like CARLA~\cite{dosovitskiy_carla_17} and AirSim~\cite{shah_airsim_2017} or procedural image generators like UnrealCV~\cite{qiu_unrealcv_2017}, BlenderProc2~\cite{denninger_blenderproc_2023} and Infinigen~\cite{raistrick_infinigen_2023} show promise to train autonomous vehicles on simulated environments with a lower Sim-to-Real Gap.
SynPhoRest~\cite{nunes_synphorest_2022, nunes_synphorest_zenodo_2022} consists of synthetic forest environments generated in the Unity game engine.
GTASynth~\cite{curnis_gtasynth_2022} provides a wide side of sequences to evaluate SLAM and registration algorithms.
The recent SORT dataset~\cite{yan_sort_2022} was generated to train an unmanned motorcycle.
At the time of writing, there is no established photorealistic synthetic dataset in field robotics.

\newlength{\forestrytablecolumnwidth}
\setlength{\forestrytablecolumnwidth}{2.7cm} % Set the height you want for your

\begin{table*}[htpb!]
\caption*{Comparison of Forestry Datasets}
\begin{tabular}{@{}>{\raggedright\arraybackslash}p{\forestrytablecolumnwidth}
>{\centering\arraybackslash}p{2cm}
>{\centering\arraybackslash}p{\forestrytablecolumnwidth}
>{\centering\arraybackslash}p{\forestrytablecolumnwidth}
>{\centering\arraybackslash}p{\forestrytablecolumnwidth}
>{\centering\arraybackslash}p{\forestrytablecolumnwidth}@{}}
\toprule
Dataset & Type & Sensors & \# Trees/Logs & \# Images & Annotation Format \\
\midrule 
BarkNet~\cite{carpentier_barknet_2018} & Real & RGB & - & 23\,000 & classification \\

SynPhoRest~\cite{nunes_synphorest_2022} & Synthetic & RGB, LiDAR, Depth & - & 3\,154 & semantic segmentation \\

CanaTree100~\cite{grondin_CanaTree100_2022}& Real & RGB & 920 & 100 & instance segmentation \\

Montmorency~\cite{tremblay_montmorency_2020}& Real & LiDAR, RGB & 1000 & - & tree species, DBH$^{*}$ \\

TimberSeg~\cite{fortin_timberseg_2022} & Real & RGB & 2\,500 & 220 & instance segmentation \\

FinnWoodlands~\cite{lagos_finn_2023}& Real & RGB, LiDAR & 2562 & 300 & panoptic segmentation \\

ForTrunkDet~\cite{dasilva_fortrunkdet_2021} & Real & RGB, Thermal & 12\,900 & 2\,895 & bounding box \\

SynthTree43k~\cite{grondin_SynthTree43k_2022} & Synthetic & RGB, Depth &  190\,000 & 43\,000 & instance segmentation\\
\bottomrule
\end{tabular}
\centering
\caption{
BarkNet is concerned with classifying trees on close-up images of their tree bark and SynPhoRest is concerned with detecting flammable material in a forest.
The majority of the forestry datasets presented here focus on detecting and segmenting tree trunks and tree logs from dense forest scenarios. \\
$^{*}$DBH is the diameter at breast height, a standard method for measuring trees in forestry.
}
\label{tab:forestry_datasets}
\end{table*}

\newlength{\forestryimagesize}
\setlength{\forestryimagesize}{5em}
\begin{figure*}[htpb!]
\begin{center}
   \rule[1ex]{.9\linewidth}{0.5pt} \\
\end{center}

\begin{center}
% BarkNet 1.0
\fcolorbox{color1}{white}
{\href{https://github.com/ulaval-damas/tree-bark-classification?tab=readme-ov-file\#barknet-10-database}{\includegraphics[height=\forestryimagesize]{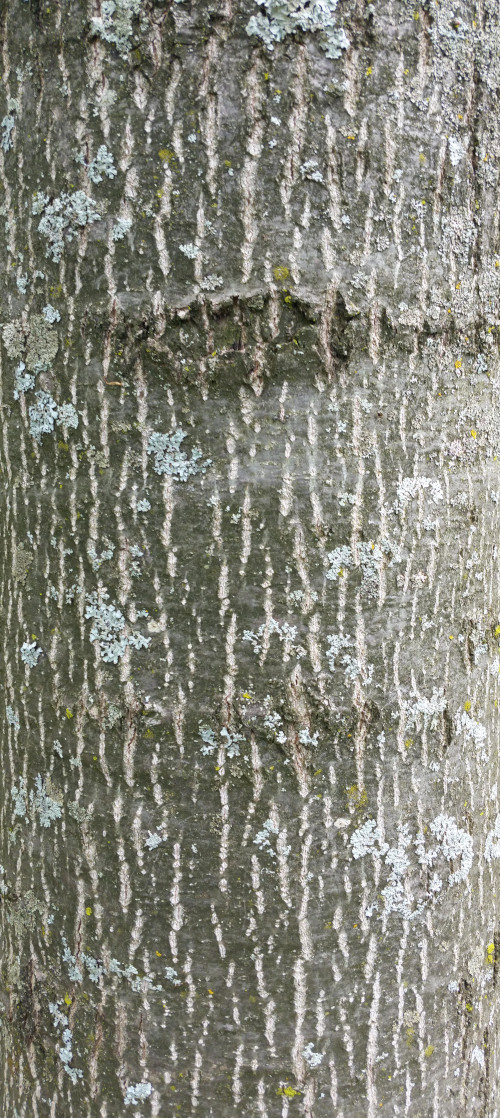}}}
% SynPhoRest
\fcolorbox{color14}{white}{\href{https://zenodo.org/records/6369446}
{\includegraphics[height=\forestryimagesize]{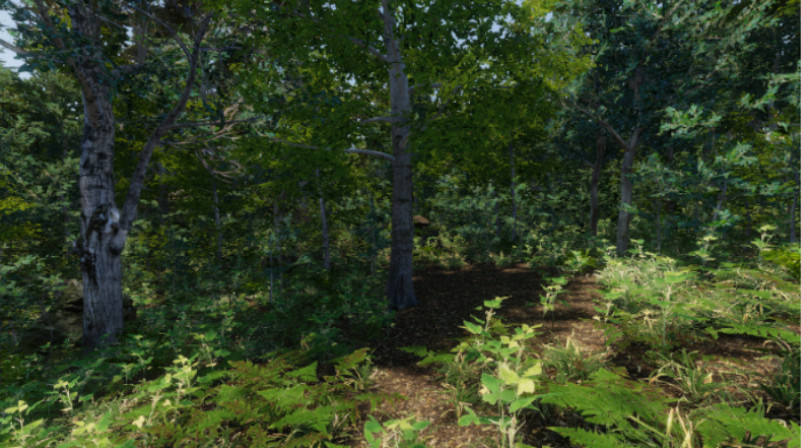}}}
% CanaTree100
\fcolorbox{color2}{white}
{\href{https://github.com/norlab-ulaval/PercepTreeV1?tab=readme-ov-file\#Datasets}{\includegraphics[height=\forestryimagesize]{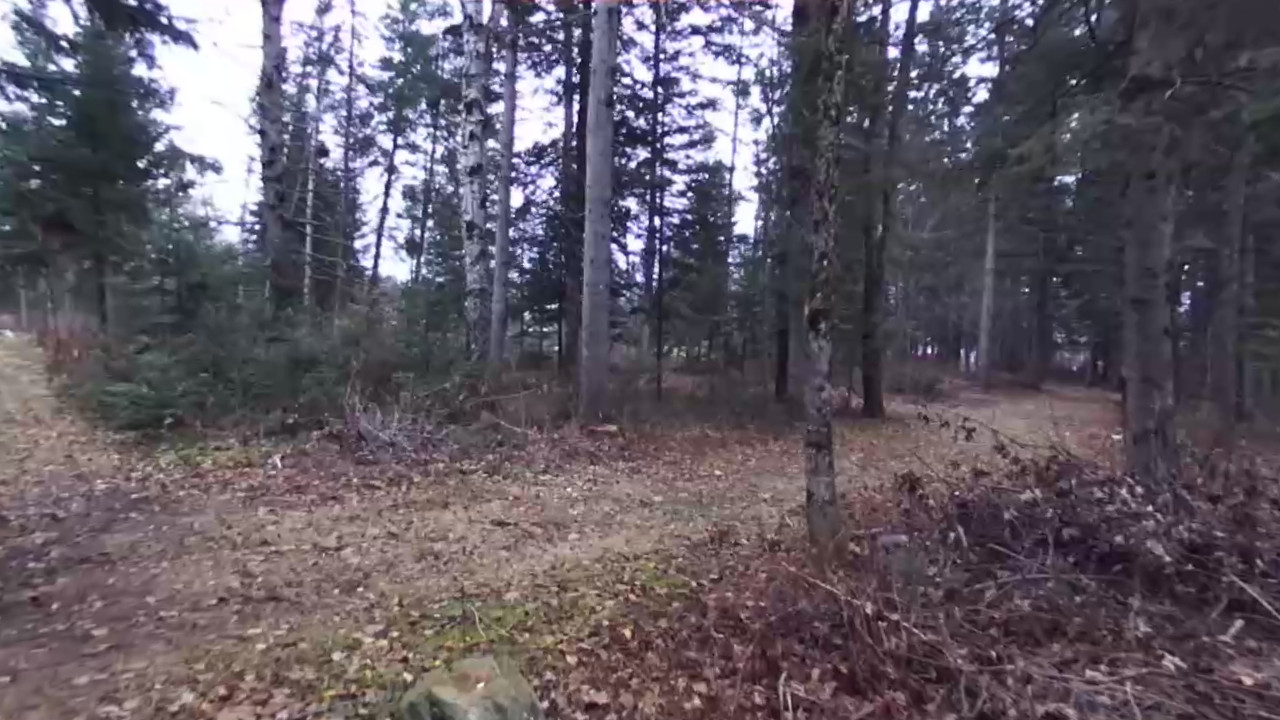}}}
% Montmorency
\fcolorbox{color9}{white}
{\href{https://norlab.ulaval.ca/research/montmorencydataset/}{\includegraphics[height=\forestryimagesize]{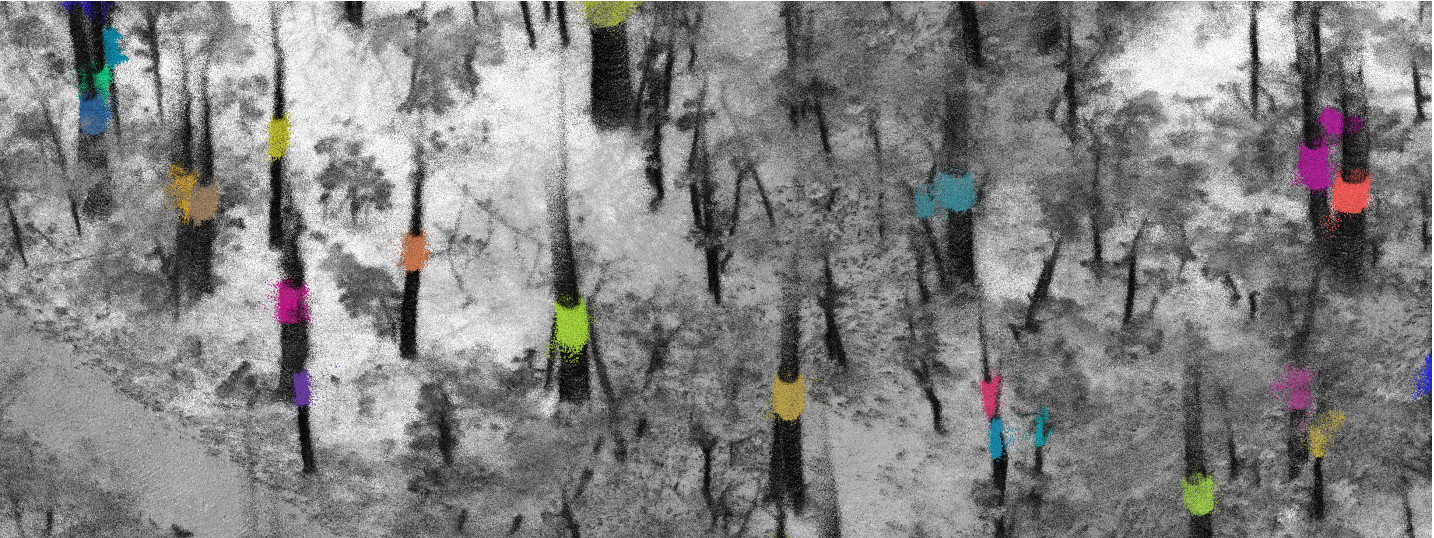}}}
% TimberSeg
\fcolorbox{color18}{white}
{\href{https://github.com/norlab-ulaval/logpiles_segmentation}{\includegraphics[height=\forestryimagesize]{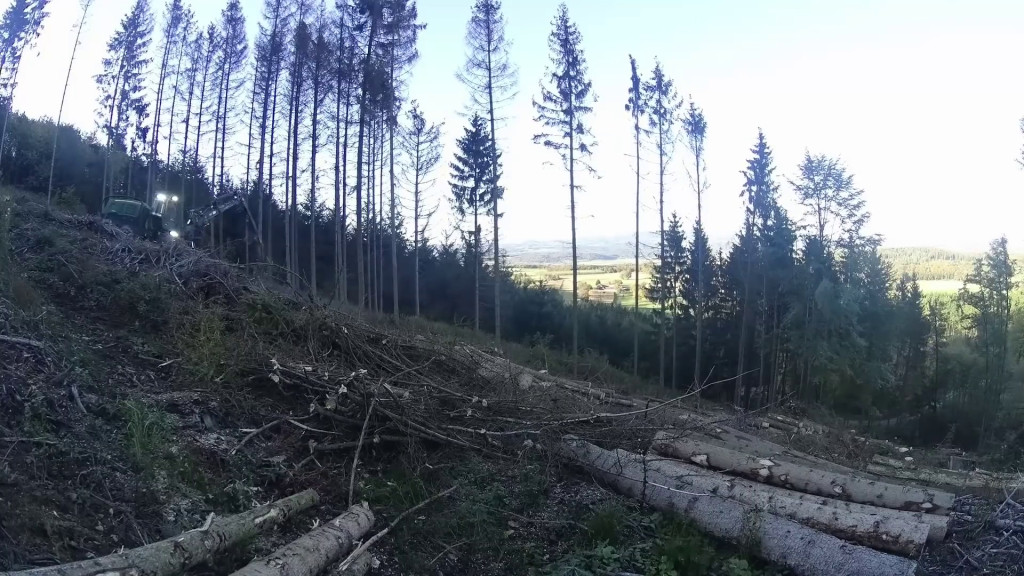}}}
% FinnWoodlands
\fcolorbox{color4}{white}
{\href{https://github.com/juanb09111/FinnForest}
{\includegraphics[height=\forestryimagesize]{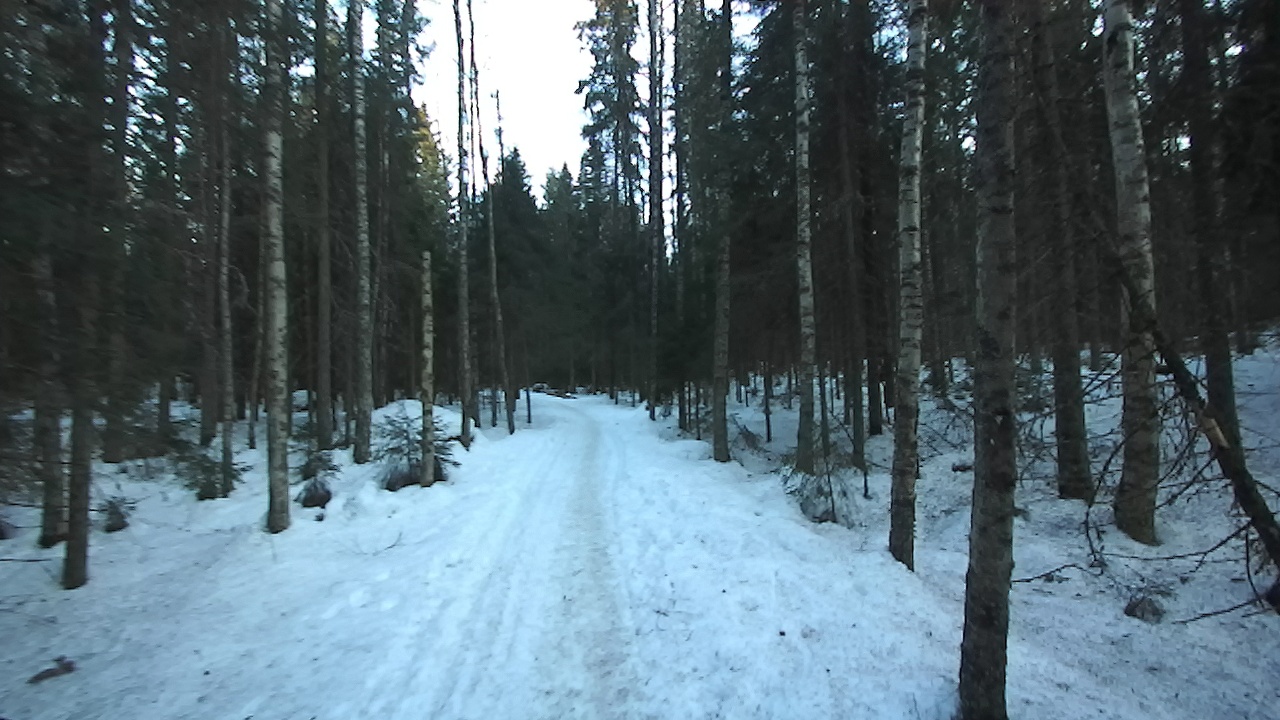}}}
% ForTrunkDet
\fcolorbox{color6}{white}
{\href{https://zenodo.org/records/5213825}
{\includegraphics[height=\forestryimagesize]{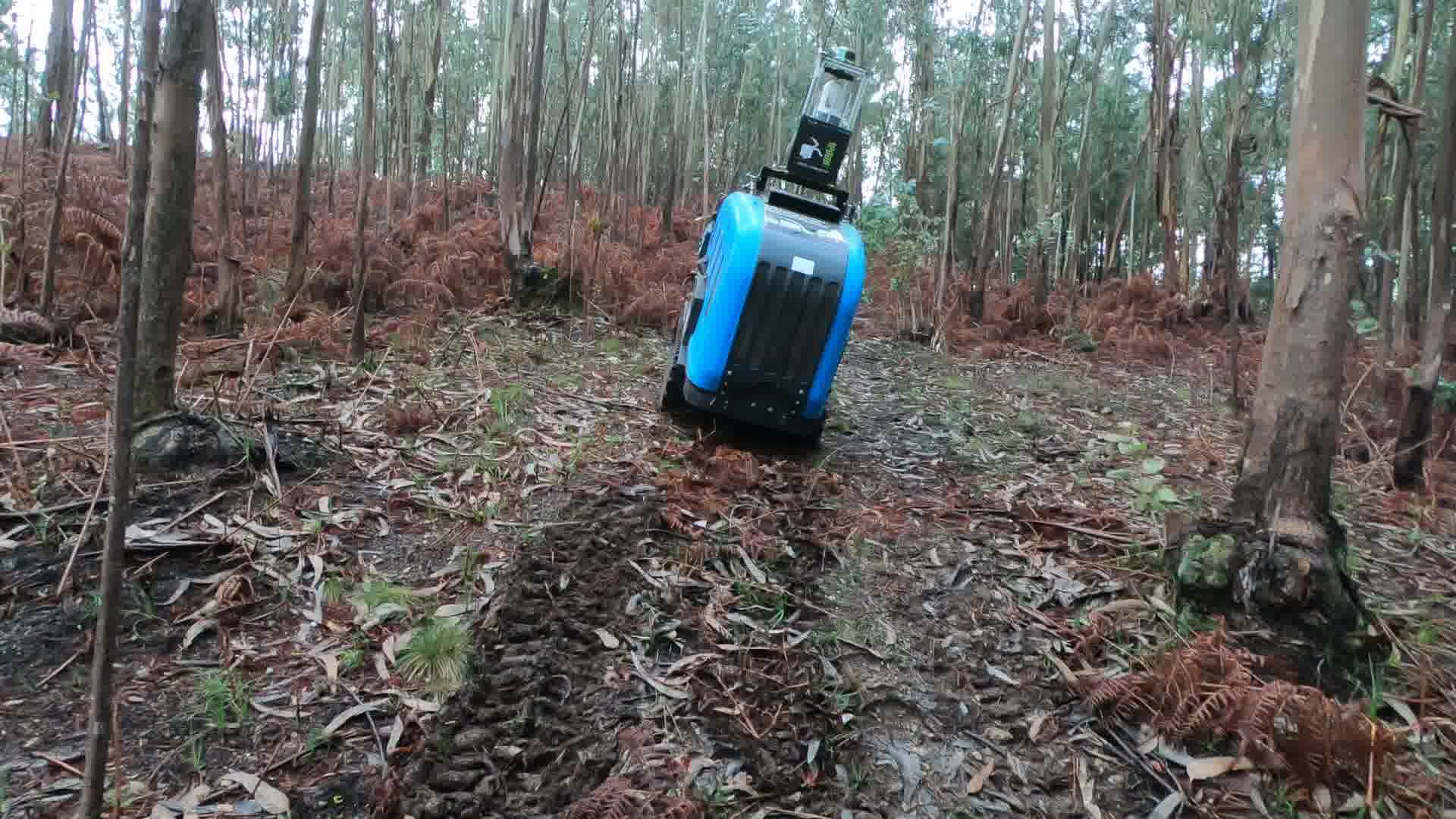}}}
% SynthTree43k
\fcolorbox{color15}{white}
{\href{https://github.com/norlab-ulaval/PercepTreeV1?tab=readme-ov-file\#Datasets}{\includegraphics[height=\forestryimagesize]{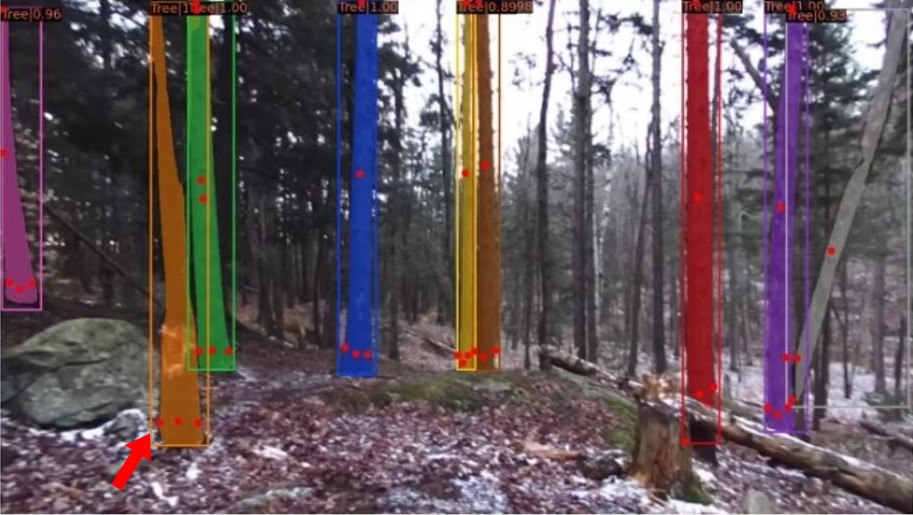}}}

\caption*{
An example image of each dataset presented in the same order as in Table~\ref{tab:forestry_datasets}.
Click on any image to open the download page for each of these datasets. Listed are
\textcolor{color1}{\textbf{BarkNet 1.0}}~\cite{carpentier_barknet_2018},
\textcolor{color14}{\textbf{SynPhoRest}}~\cite{nunes_synphorest_2022},
\textcolor{color2}{\textbf{CanaTree100}}~\cite{grondin_CanaTree100_2022},
\textcolor{color9}{\textbf{Montmorency}}~\cite{tremblay_montmorency_2020},
\textcolor{color18}{\textbf{TimberSeg}}~\cite{fortin_timberseg_2022},
\textcolor{color6}{\textbf{FinnWoodlands}}~\cite{lagos_finn_2023},
\textcolor{color6}{\textbf{ForTrunkDet}}~\cite{dasilva_fortrunkdet_2021},
\textcolor{color15}{\textbf{SynthTree43k}}~\cite{grondin_SynthTree43k_2022}.
}

\end{center}

\begin{center}
    \rule[1ex]{.9\linewidth}{0.5pt} \\
\end{center}

\end{figure*}

\subsection{Terrain Datasets}
\label{subsec:terrain_datasets}

A common perception task in field robotics, is the detection of the road when navigating along a forest service road or a field path.
These paths often require specialized approaches compared to the road tracking in urban scenarios.
For one, poor GNSS conditions in forested areas prevent heavy reliance on a global localization for tracking the road and the lack of road markings like on paved roads prevent the easy detection of the edges of the road surface~\cite{forkel_roadtracking_2021, forkel_roadtracking_2022}. \\
In Table~\ref{tab:terrain_datasets}, we compare different datasets released for the specific task of path following. 
TrailNet~\cite{hoveidar_trailnet_2018} is one of the first datasets to analyze its approach for different road surface types and release the recorded camera data publicly.
ORFD~\cite{min_orfd_2022} makes use of the calculated surface normals in the LiDAR point cloud data to train a Transformer encoder combined with camera images to detect free space. 
The authors provide ground truth annotations of the free space to allow for quantitative evaluations. \\
Rally Estonia~\cite{tampuu_estoniadriving_2023} released over \SI{750}{\kilo\meter} of recorded wide angle camera and LiDAR data of driving along a gravel road with low traffic in Estonia. 
The same road was recorded during different seasons and also during adverse weather conditions like heavy rain.
Rally Estonia was originally intended for End-to-End driving models that predict lateral control (steering) based on the input image and LiDAR data.
The original data was recorded with a human driver.
Combined with the VISTA~\cite{amini_vista2_2022} simulator, the recorded sensor data can be reprojected to simulate the viewpoint of the desired trajectory during model evaluation. 
The evaluation metric is then based on the deviation of the End-to-End driving model from the human-driven center-line. \\
Vale~\cite{hosseinpoor_vale_2021} is very similar to CaT~\cite{sharma_CaT_2022} in terms of laying a focus on trafficability of different surfaces.
The dataset in Vale was recorded from an aerial perspective with a drone.
The image data is then annotated based on different levels of traversability that depend on the maximum height difference within a certain surface region.
These levels of traversability are then associated with typical methods of robot locomotion such as \textit{Wheeled}, \textit{Tracked} and \textit{Legged}.
This allows for a more direct interpretation of the surface as a navigable space compared to classifying surfaces based on their visual appearance.
Overall, each terrain dataset was specifically developed for one application, which can vary from supervised learning for perception to evaluating End-to-End models.
We believe this is due to the limited annotation effort necessary to segment free space and road surfaces.

\subsection{Forestry Datasets}

In field robotics for forestry, autonomous systems can automate the management of trees and logs within a forested area. 
The datasets available either focus on detecting the individual tree trunks in a scene or on detecting the tree logs of a recently harvested forest.
BarkNet 1.0~\cite{carpentier_barknet_2018} contains 23\,000 images of tree barks that are classified into 23 different tree species for automated tree species identification.
ForTrunkDet~\cite{dasilva_fortrunkdet_2021} contains 2\,895 bounding box annotations of tree trunks in color images and in thermal images as well.
For the particular task of log grasping, a bounding box detection is not sufficient.
Datasets like TimberSeg 1.0~\cite{fortin_timberseg_2022} provide instance segmentations of logs for autonmous log grasping. 
TimberSeg 1.0 consists of 220 images with 2500 instances of wood logs in total.
CanaTree100~\cite{grondin_CanaTree100_2022} is a dataset for tree instance segmentation with 100 images and over 920 annotated trees collected in Canadian forests.
SynthTree43k~\cite{grondin_SynthTree43k_2022} consists of 43\,000 synthetic images with 190\,000 annotated trees in an environment created in the Unity game engine.
As for many outdoor applications, the geographical location of the dataset recording has a great effect on the visual appearance of the environment.
The trees of the forest in Portugal in ForTrunkDet are different in size and diameter than the trees seen in CanaTree100.
Since this perception task of tree instance segmentation requires a low amount of annotation effort, one could imagine a large scale tree instance dataset that covers many areas on the globe similar to the Mapillary Vistas dataset~\cite{neuhold_mapillary_2017}.

\subsection{Place Recognition}

Place recognition can be of use in localization frameworks in different forms.
The recognition of an already observed place within the same sequence (intra-sequence) can support the loop closure detection. 
Also the recognition of places between two sequences (inter-sequence) to re-localize the robot platform after a shutdown or a long period of dropped GNSS coverage can be solved using visual place recognition methods.
For a robust evaluation of place recognition methods, the datasets require long sequences of data recording across a large spatial scale with many loops and reverse revists of parts of the sequence.
For structured urban environments, datasets like Oxford RobotCar~\cite{maddern_oxfordrobotcar_2017} has many of these features and covers a \SI{1000}{\kilo\meter} of driving.
The closest equivalent for unstructured outdoor environments is the recent Wild-Places~\cite{knights_wildplaces_2023} dataset that covers eight sequences recorded in national parks in Brisbane, Australia.
Wild-Places was recorded with a handheld sensor payload consisting of a rotating Velodyne VLP-16 LiDAR scanner with four cameras to offer a $360^\circ$ surround view. \\
TB-Places~\cite{leyvavallina_tbplaces_2019, leyvavallina_place_2019} consists of camera images taken by a robot platform in different gardens across four recording sessions with varying light conditions.
TB-Places differs from Wild-Places by not containing any LiDAR point cloud data and focusing on the garden environment with a small spatial scale. \\
The VPAIR~\cite{schleiss_vpair_2022} is an aerial visual place recognition dataset that covers \SI{107}{\kilo\meter} on a flight trajectory across urban, rural, forestry and agricultural areas. 
This recorded aerial image data is combined with publicly available surface model data in form of point clouds with \SI{0.5}{\meter} accuracy. 
The use of VPAIR for field robotics requires methods that can use aerial image data to recognize the robot's map position based on its ground viewpoint. \\

\subsection{Robot Platforms}

With open-source robotics middlewares like ROS~\cite{quigley_ros1_2009} gaining traction, it has become increasingly possible to also release the raw sensor in addition to the post-processed data used for the evaluations.
The ROS bag format does not just hold the raw sensor data, but it can also hold information about the frames of references and their relative transformations in the same file.
The arrival of the sensor data can also be replayed from a ROS bag.
More recent outdoor datasets like RELLIS-3D~\cite{jiang_rellis-3d_2021} provide URDF models of their robot platform that can be used in combination with the ROS bag data to use to evaluate perception tasks like semantic mapping or path detection. \\
Projects like TartanDrive 2.0~\cite{sivaprakasam_tartandrive2_2024} aim at providing a framework around their robot platform, that can be used in combination with released data recordings to train models using self-supervised approaches. \\
MAVS~\cite{sharma_CaSSeD_2022} relies on physics-based sensor simulations, digital terrain generation and a vehicle simulator to simulate autonomous vehicles in digital terrain.
With the release of raw sensor data comes the difficulty of providing the correct software packages, such as drivers for each sensor, to ensure that even older recordings can be replayed. 
Here the concept of linking the raw sensor data with a particular version of the perception platform using version control can ensure the reproducibility of the raw sensor data~\cite{vizzo_ipb_2023}. \\
Overall, the release of more of the auxiliary sensor data apart of the annotated sensor opens up the possibility to evaluate methods that use more complex input data like multimodal neural networks~\cite{valada_ssma_2019} or using sequential frames as input~\cite{chen_mos_2022}.  
The migration to ROS 2~\cite{macenski_ros2_2022} appears to still be ongoing for most robotic research groups with none of the mentioned datasets in this survey releasing their raw sensor data in the ROS 2 bag format.

% \subsection{Urban Dataset with Unstructured Outdoor Environments}
% here I could mention IDD and Mapillary Vistas

%\section{Opportunities with Autolabeling}
% Link to the web demo of SEEM
% http://semantic-sam.xyzou.net:6090/

\section{Discussion and Conclusion}

In this survey, we have presented many of the datasets available for different perception tasks in field robotics in unstructured outdoor environments.
In the following we will highlight some general observations that arise when reviewing many field robotics datasets.\\
For one, there is no dataset that is being used across many perception tasks, but rather a smaller set of specialized datatsets for each particular perception task. 
This is due to varying requirements on the datasets for specific tasks like End-to-End methods for path following (see Rally Estonia~\cite{tampuu_estoniadriving_2023}), semantic segmentation (see RELLIS-3D~\cite{jiang_rellis-3d_2021}) or visual place recognition (see Wild-Places~\cite{knights_wildplaces_2023}). \\
This differs from the research ecosystem around autonomous driving in urban environments, where datasets like KITTI~\cite{geiger_vision_2013} to a greater extent or CityScapes~\cite{cordts_cityscapes_2016} to a lesser extent serve as benchmark for many different tasks.
The advantage of broader datasets like KITTI is, that they draw enough attention to warrant the maintenance of a public evaluation server, which greatly improves the ease of quantitatively comparing methods for the same perception task.
The issue that arises in the field robotics ecosystem is that the effort to setup and maintain evaluation servers is spread across many smaller research communities, where the barrier of entry to setup an evaluation server is still too high. 
Open source platforms like CodaLab~\cite{pavao_codalab_2023} are lowering this barrier, but only very few competitions are maintained in the long term. Commerical platforms like Kaggle~\cite{kaggle} can draw many participants with cash prizes, but this would add an additional financial burden to setting up an evaluation server. \\
For the perception task of 2D semantic segmentation, there are many public research datasets that have been released in recent years (see Table~\ref{tab:2d_segmentation_datasets}).
A composite dataset similar to MSeg~\cite{lambert_mseg_2020} can be constructed to train a robust semantic segmentation system that can applied across different sensor setups in unstructured outdoor environements.
This would require a unified taxonomy to combine the different levels of annotation granularity across these datasets.
We believe, efforts to create ontologies such as ATLAS~\cite{smith_atlas_2022}, that can map labels across different segmentation datasets, can ease the conversion of datasets to a unified semantic representation. 
These ontologies can also be applied to annotated LiDAR data, but as Table~\ref{tab:3d_segmentation_datasets} has shown, only very few datasets are publicly available in this research area.\\ 

What this survey revealed is that the area of field robotics has grown in recent years.
Viewing the research area through the lens of public research datasets, we observe many new datasets that were published in the last 5 years. \\
Even though this survey did not take a closer look into the datasets published in agricultural robotics, we believe this survey to be valuable for many practitioners that want to get a general overview of research datasets available for field robotics. \\ 
We look forward to foundation models like SEEM~\cite{zou_seem_2023} to reduce the human labor required in data annotation even further to open up the possibility of deploying supervised learning algorithms in highly specialized domains found in field robotics. \\

\balance
% Generated by IEEEtran.bst, version: 1.14 (2015/08/26)

%\bibliographystyle{IEEEtran}
%\bibliography{bib/IEEEfull,bib/additional_full,bib/literature}
% that's all folks
\end{document}